\documentclass{article}
\usepackage[preprint]{tmlr}
\usepackage{comment}
\usepackage{amsmath,amssymb,amsfonts}
\newcommand{\Description}[1]{} %

\usepackage{booktabs}
\usepackage{algorithm}
\usepackage[noEnd]{algpseudocodex}
\renewcommand{\algorithmicrequire}{\textbf{Input:}}
\renewcommand{\algorithmicensure}{\textbf{Output:}}
\usepackage{graphicx}
\usepackage{subcaption}
\usepackage{xcolor}
\usepackage{tikz}
\usetikzlibrary{arrows.meta,positioning,shapes.geometric,fit,backgrounds,calc}

\newcommand{\system}{FunnelAL}
\begin{document}

\title{FunnelAL: Retrieve-then-Rank Active Learning for Single-Class Discovery}

\author{\name Reihaneh Rostami \email reihaneh@raiclabs.com \\
\addr RAIC Labs, USA
\AND
\name Brian Goodwin \email brian@raiclabs.com \\
\addr RAIC Labs, USA}

\maketitle

\begin{abstract}
We present \system{}, a retrieve-then-rank active learning system for single-class discovery, which adapts the multi-stage funnel architecture of industrial recommender systems to data annotation. Large-scale supervised learning faces two challenges: efficiently finding relevant samples in a massive corpus, and distinguishing true positives from visually confusable negatives when embeddings do not cleanly separate classes. Most active learning scores the unlabeled pool with one fixed acquisition function.  Retrieval-based methods narrow the pool first, but still select with a single criterion. \system{} decomposes the problem into cascaded stages. Starting from a single positive and negative example, the system iterates through: (1)~embedding-based retrieval scoring that narrows the corpus to a manageable candidate set; (2)~a ranking stage that exploits a learned ranker (RankNet) and, once batch precision falls, mixes in committee-based exploration (QBC); and (3)~feedback from the annotator's labels that refines both stages in subsequent iterations.  We evaluate on five datasets, three benchmarks and two large corpora of up to 2.7 million images, each with two different pretrained encoders, DINOv2 and CLIP.  With clean labels and DINOv2, \system{} has the best mean rank over the five datasets, and with CLIP it shares the best rank with its ablation Funnel-RankNet.  On the three benchmarks it discovers positives significantly faster than the classical and uncertainty-based baselines and than PF-MA, and no significant difference from PF-MA's final F1 was detected.  With DINOv2, no significant difference from SEALS's active-search mode (most-likely-positive selection) was detected on most datasets.  With CLIP, \system{} discovers significantly faster than it on three of the five.  On iNaturalist's working set of 2.15 million images \system{} selects eight times faster than the neighborhood baselines as we implemented them (SEALS in both its modes and AnchorAL, which select only among the nearest neighbors of labeled items). There, PF-MA and uncertainty sampling, which label items near the decision boundary across the whole pool, reach a higher F1 under our evaluation protocol.  Under 10\% annotator label noise it has a higher F1 than every baseline on three of the five datasets with DINOv2 and on all five with CLIP, and SEALS's active-search mode is more robust at higher rates.  Our work provides a concrete bridge between multi-stage recommender systems and active learning, offering both a perspective on annotation-as-recommendation and a practical system for label-efficient classification across diverse domains.
\end{abstract}

\section{Introduction}\label{sec:intro}

Supervised learning at scale requires labeled data, and obtaining it remains one of the most persistent bottlenecks in machine learning practice.  The problem involves two challenges.  First, \emph{scale}: corpora are often large, and the cost of scoring every candidate at every iteration grows with both the corpus size and the complexity of the scoring model.  Efficiently narrowing the pool to relevant candidates becomes critical as corpus size grows~\citep{settles2012active}.  Second, \emph{class confusability}: even after relevant candidates have been found, distinguishing the true target from visually similar distractors is difficult~\citep{wah2011caltech, tuia2011survey} when the available feature representations do not cleanly separate classes. This challenge can arise when the features lack the discriminative power needed for fine-grained distinctions or when the annotation domain lies far from the training distribution of the embedding model~\citep{oquab2024dinov2, gupte2024revisiting}.  These two challenges call for different solutions.  The question is therefore not merely \emph{which} samples to label, but \emph{how to structure the selection process} so that retrieval efficiently narrows the pool to relevant candidates and ranking surfaces the samples most effective for refining the decision boundary as labels accumulate.

Active learning provides a principled answer: select the samples from which the model learns the most~\citep{settles2012active}. Query-by-Committee (QBC)~\citep{seung1992query, freund1992information} targets disagreement among an ensemble, and uncertainty sampling targets the model's least-confident predictions. These strategies have been shown to dramatically reduce annotation budgets~\citep{gilad2005query, settles2012active}. Yet conventional active learning treats sample selection as a single-stage process: one acquisition function scores the entire unlabeled pool per iteration. This design carries three practical limitations. First, it scales poorly with corpus size. Scoring every candidate in a large corpus at each iteration is computationally expensive and often unnecessary.  Retrieval-based methods address this directly by restricting the candidate pool to a similarity neighborhood of the labeled set~\citep{coleman2022similarity, lesci-vlachos-2024-anchoral}.  Second, it commits to a single, fixed acquisition function.  Methods that choose among acquisition functions online relax this~\citep{baram2004online, hsu2015albl}, adapting which single function runs rather than separating the two concerns into distinct stages.  The balance between exploiting known positives and exploring uncertain regions cannot adapt as the session progresses, and the annotator is given no leverage over it, remaining a passive oracle~\citep{lee2020empowering} rather than a participant whose domain intuition could guide the search.    Third, classical active learning typically assumes a multi-class setting in which labels for all classes are collected simultaneously and cross-class information aids selection~\citep{settles2012active}.  A recent line of work instead targets rare-concept discovery, framing it largely as binary or one-vs-rest problems~\citep{coleman2022similarity, lesci-vlachos-2024-anchoral, zhang2022galaxy}.  Yet in many practical scenarios, an analyst needs to find instances of a \emph{single} class of interest, e.g., a specific vehicle type in satellite imagery~\citep{tuia2011survey}, a particular species in an ecological survey~\citep{norouzzadeh2018automatically}, or a target lesion in medical scans~\citep{tschandl2018ham10000}, starting from as little as one reference example and with no taxonomy of other classes available.  This single-class discovery setting demands methods that can start from a single labeled example and cannot lean on labels from other classes.

Recommender systems face a structurally similar problem: selecting a few relevant items from massive catalogs through iterative user feedback. The field has developed architectural patterns that annotation systems have yet to exploit. The multi-stage \emph{funnel} is the dominant paradigm in industrial recommendation: a cheap retrieval model narrows millions of candidates to hundreds, then a more expressive ranker scores and re-ranks the shortlist~\citep{covington2016deep, ma2018entire, qin2022rankflow}. This cascaded design reduces computational cost while preserving ranking quality.  The expensive ranker scores only the shortlist, never the full catalog. Interactive and conversational recommender systems go further, refining their outputs over multi-turn dialogues with the user through estimate-act-reflect cycles~\citep{lei2020estimation, gao2021advances, antognini2021multi}. And bandit-based recommendation formalizes the tension between showing items the system is confident about (exploitation) and probing uncertain regions to improve future predictions (exploration)~\citep{li2010contextual, chapelle2011empirical, su2024exploration}. Despite these parallels, the connection between recommendation funnels and active annotation has remained largely conceptual. Prior work has noted the overlap between active learning and recommendation~\citep{rubens2015active, elahi2016survey}, and recent methods apply active learning to training rankers~\citep{wang2022active} or cast acquisition itself as a learned ranking~\citep{ding2024learning}, but, to the best of our knowledge, no prior annotation system decomposes sample selection into an explicit retrieval stage followed by a two-arm ranking stage with an adaptive exploitation-to-exploration transition.

We bridge this gap with \system{}, which treats data annotation as a multi-stage recommendation problem.  \system{} is designed for \emph{single-class discovery}: the annotator has one class of interest and wants to find its instances in a large corpus, starting from just one positive and one negative seed, with no multi-class taxonomy and no cross-class signal to aid selection.  In this framing, unlabeled samples are \emph{items} to be recommended, the annotator is the \emph{user} whose preferences the system must learn, and each annotation round is a \emph{session} in which the system presents a ranked list and receives feedback.  The system iterates through three stages that mirror the industrial recommendation pipeline:

\begin{itemize}
    \item \textbf{Retrieval.} Embedding-based similarity search narrows the full corpus to a candidate set~$\mathcal{C}$, analogous to the candidate generation stage in large-scale recommender systems~\citep{covington2016deep}.
    \item \textbf{Ranking.} A two-arm ranking stage selects each batch: a \emph{RankNet}-based ranker~\citep{burges2005learning} that exploits current knowledge to surface likely positives, and a \emph{Query-by-Committee} ranker~\citep{seung1992query} that explores uncertain regions of the decision boundary. A precision trigger shifts the stage from pure exploitation to a fixed blend of both arms once the precision of the exploitation batches falls below a threshold, echoing the bandit explore-exploit trade-off~\citep{li2010contextual}.  As a design affordance, the switch can also be placed under direct annotator control.
    \item \textbf{Feedback.} The annotator's labels update the positive and negative example sets, refining both retrieval and ranking for the next iteration, echoing the iterative feedback loops in conversational recommendation~\citep{lei2020estimation}.
\end{itemize}

\noindent At each iteration, a classifier trained on the accumulated labels and presumed negatives is evaluated on a held-out test set to measure how effectively the recommended annotations capture the target class.

Our contributions are as follows:
\begin{enumerate}
    \item \textbf{Formalization.} We formalize data annotation as a multi-stage funnel recommendation problem, connecting two largely separate research communities, i.e., recommender systems and active learning, and opening annotation to the toolkit of funnel optimization and interactive recommendation.
    \item \textbf{System.} We propose \system{}, to our knowledge the first annotation system that couples an explicit retrieval stage with a two-arm (exploit/explore) ranking stage and an adaptive transition between the arms, starting from a single positive and negative example.
    \item \textbf{Adaptive exploration-exploitation.} Active learning that adapts its acquisition strategy exists, typically by selecting among candidate strategies~\citep{baram2004online, hsu2015albl}.  \system{} instead makes the explore-exploit balance adaptive within a single ranking stage: a precision trigger defers exploration until the precision of the exploitation batches falls.  On the three benchmarks with DINOv2 and clean labels, our experiments show that exploring from the first rounds slows discovery.  No significant difference was detected between the precision trigger and a switch at round 5 or 8, or, with the batch fixed at 20, a switch after 30 positives (Section~\ref{sec:ablation}).  With DINOv2, no significant difference was detected on the benchmarks between \system{} and an ablation that never explores, Funnel-RankNet, and with CLIP Funnel-RankNet is slightly ahead on some discovery columns (Section~\ref{sec:results}, Appendix~\ref{app:clip}).  On iNaturalist with DINOv2 and clean labels, exploration trades discovery for classifier quality: Funnel-RankNet discovers slightly faster and \system{} has the higher AULC-F1, both significant.  The balance is also exposed as a configurable affordance for direct annotator control.  Our experiments evaluate the automated hybrid policy, and a user study of manual control is left to future work.
\end{enumerate}

We evaluate \system{} on three benchmarks, CUB-200-2011~\citep{wah2011caltech} (200~bird species), FGVC-Aircraft~\citep{maji2013fgvc} (90~aircraft classes) and UC Merced Land Use~\citep{yang2010ucmerced} (21~aerial land-use classes), and on two large corpora, iNaturalist 2021~\citep{vanhorn2021inat} (2.7 million photographs) and fMoW~\citep{christie2018fmow} (83,412 sites of its training split, one image each), each with two encoders.  With clean labels and DINOv2 \system{} has the best mean rank over the five datasets and discovers positives faster than every external method except SEALS's active-search mode, from which no significant difference was detected on most datasets.  At scale its ranker scores a bounded number of candidates, and as we implemented the methods it selects faster than the neighborhood baselines.  On the largest corpus, PF-MA and uncertainty sampling reach the higher F1.
    
The remainder of this paper is organized as follows. Section~\ref{sec:related} reviews related work across multi-stage recommendation, interactive recommendation, bandit-based exploration, active learning, and annotation systems. Section~\ref{sec:method} describes the \system{} framework in detail. Section~\ref{sec:experiments} presents the experimental setup, baselines, and results, including selection cost, the timing of exploration, label noise, and a second encoder. Section~\ref{sec:discussion} analyzes results across datasets and discusses limitations and future directions, and Section~\ref{sec:conclusion} concludes.  The appendices report retrieval breadth on iNaturalist, the evaluation classifier, the trigger window, retrieval from the unlabeled pool, and GAL and PF-MA's probability map.  They also report the ablations at 10\% noise, the sensitivity of \system{}'s settings, the second encoder in full, and the target classes of the large corpora.

\section{Related Work}\label{sec:related}

Selecting which samples to present to a human labeler sits at the intersection of recommender systems, active learning, and human-in-the-loop annotation.

\noindent\textbf{Interactive Multi-Stage Recommender Systems.} The retrieve-then-rank funnel is the dominant architecture in industrial recommendation, established by \citet{covington2016deep} for YouTube and extended to multi-task engagement modeling along the impression-click-conversion funnel~\citep{ma2018entire}, algorithm-system co-design for pre-ranking~\citep{wang2020cold}, joint stage training~\citep{qin2022rankflow}, and training the whole cascade end to end as one network~\citep{wang2025lcron}.  Unlike production recommendation, where the funnel is optimized for engagement metrics, our funnel is optimized for annotation efficiency by maximizing what each human interaction teaches the system.

Interactive and conversational recommender systems refine their outputs through multi-turn user feedback~\citep{gao2021advances}, using estimate-act-reflect cycles~\citep{lei2020estimation}, structured critiques~\citep{antognini2021multi}, or open-ended preference elicitation with language models~\citep{li2025gate}.  \system{} shares this iterative feedback-driven structure but uses binary annotation labels rather than natural-language dialogue.

\noindent\textbf{Exploration and Exploitation in Recommendation.} The tension between exploitation and exploration is formalized by contextual bandits~\citep{li2010contextual}, with alternatives such as Thompson Sampling~\citep{chapelle2011empirical}, collaborative filtering~\citep{gentile2014online, li2016collaborative}, and neural representations~\citep{zhou2020neural, su2023nonlinear}. Production deployments confirm that exploration yields measurable long-term value~\citep{su2024exploration, su2024multitask}. 

These methods all operate within a single selection stage.  The bandit policy \emph{is} the acquisition function.  In conventional active learning, the explore-exploit balance is similarly folded into the choice of acquisition function rather than separated into distinct stages.  \system{} decouples them by placing exploitation and exploration in distinct ranking components and introducing a precision-triggered policy that transitions between the two as annotation progresses, as detailed in Section~\ref{sec:method:adaptive}.

\noindent\textbf{Active Learning and Learning to Rank.} Active learning reduces annotation cost by selecting the samples expected to be most informative~\citep{settles2012active}.  Query-by-Committee~\citep{seung1992query, freund1992information} is among the oldest such strategies, later made computationally practical through kernelized version-space sampling~\citep{gilad2005query}.  BADGE~\citep{ash2020badge} combines uncertainty and diversity in gradient space, and TypiClust~\citep{hacohen2022typiclust} identifies a phase transition from typical to atypical queries as budgets grow.  ProbCover~\citep{yehuda2022probcover} maximizes embedding-space coverage, relying on high-quality representations where spatial proximity implies label agreement.  \system{}'s ranking stage instead learns pairwise preferences directly from accumulated labels, so it does not rely on embedding geometry alone, although the retrieval stage does.  DropQuery~\citep{gupte2024revisiting} operates on foundation-model embeddings, applying dropout-based uncertainty filtering to the unlabeled pool followed by clustering-based diversity selection on the resulting candidates. Evaluated on natural and out-of-domain biomedical images, it does not use a similarity-based retrieval stage before uncertainty scoring.  All these methods operate as monolithic acquisition functions.  None decompose selection into retrieval and ranking.  On the ranking side, RankNet~\citep{burges2005learning} and its successors LambdaRank/LambdaMART~\citep{burges2010ranknet} learn pairwise preferences, and recent work connects active learning to ranking~\citep{wang2022active} and bilevel optimization~\citep{ding2024learning}. 

The connection between active learning and recommendation has been noted in the literature.  Both must select items to present to a user under uncertainty~\citep{rubens2015active, elahi2016survey}.  However, this connection has remained at the conceptual level.  To the best of our knowledge, no prior annotation system couples an explicit retrieval stage with a two-arm ranking stage and an adaptive transition between the arms.

\noindent\textbf{Human-in-the-Loop Annotation Systems.} Recent annotation systems treat the human as more than a passive oracle by integrating active learning into crowd annotation~\citep{lin2019alpacatag}, surveying data- and model-centric paradigms~\citep{wu2022hitl}, and leveraging LLMs for pre-annotation~\citep{tan2024llmannotation}.  \system{} operates in this tradition.  The annotator provides binary feedback on each recommended sample, and the system refines its retrieval and ranking in response.  Beyond labeling, \system{}'s architecture also allows the annotator to directly control the explore-exploit balance by choosing between RankNet and QBC ranking at each iteration, though our experiments evaluate only the automated precision-triggered policy.

\noindent\textbf{Scalable Active Learning and Single-Class Discovery.} Standard pool-based active learning scores every unlabeled sample per iteration.   SEALS~\citep{coleman2022similarity} targets rare-concept discovery at web scale, framing each concept as an independent binary classification problem and restricting the candidate pool to a similarity-based neighborhood of the labeled set, matching full-pool mean average precision and recall at a fraction of the selection cost.  Its published selection strategies include max-entropy uncertainty and a most-likely-positive rule for active search.  AnchorAL~\citep{lesci-vlachos-2024-anchoral} targets large, imbalanced pools, evaluated mainly on binary rare-class tasks.  It dynamically selects diverse class-specific anchors from the labeled set and retrieves their nearest neighbors to form a fixed-size subpool at each iteration, promoting exploration and class balance while keeping computational cost constant regardless of pool size.  \citet{citovsky2021batch} further scale batch-mode active learning to batch sizes of 100K--1M by combining margin-based uncertainty with hierarchical agglomerative clustering for diversity.  GALAXY~\citep{zhang2022galaxy} directly targets the rare-class regime by blending graph-based active learning with deep learning, building a per-class one-vs-rest linear graph over the model's confidence margins.  It performs a refined uncertainty sampling that gathers more class-balanced batches than vanilla uncertainty sampling.    All of these methods apply a single, fixed acquisition function after their pre-filtering step.  Adaptive selection among acquisition strategies has itself been studied: COMB~\citep{baram2004online} chooses among active learning algorithms online with a multi-armed bandit whose reward estimates classification progress, and Active Learning by Learning~\citep{hsu2015albl} formulates the choice as an exploration-exploitation problem over strategies with an importance-weighted reward.  \system{}'s transition differs in kind from these: a one-way, precision-triggered phase change between an exploitation arm and an exploration arm inside a retrieval-restricted candidate set, rather than bandit-style reselection among full-pool strategies.  Two recent works share \system{}'s interactive single-class discovery setting most closely: GAL~\citep{bar2024gal} introduces impact-based acquisition functions with greedy batch selection for interactive image retrieval starting from few labeled samples, and PF-MA~\citep{zaher2026pfma} proposes a positivity-biased uncertainty criterion for discovering rare categories under extreme class imbalance.    None of the above couples an explicit retrieval stage with a two-arm ranking stage and a precision-triggered one-way transition between the arms.  That specific conjunction, rather than any individual component, is what \system{} contributes.  Active learning has also shown particular value in specialized domains where annotations are expensive, including remote sensing~\citep{tuia2011survey, lenczner2022dial, robinson2020human}, medical imaging~\citep{tschandl2018ham10000}, and fine-grained recognition~\citep{wah2011caltech}, but these systems likewise do not adopt a multi-stage funnel architecture.

Among these methods, SEALS (in two of its published selection modes), AnchorAL and PF-MA serve as experimental baselines in Section~\ref{sec:experiments}, and GAL is compared in Appendix~\ref{app:gal} on the three benchmarks, under the same embeddings, train/test split, budget, cold start (one positive and one negative seed), and evaluation classifier as \system{}, its ablations, and the classical baselines.  SEALS and AnchorAL are the most architecturally comparable to \system{}.  All three use embedding-based similarity search to pre-filter the pool before applying a selection criterion, so the comparison centers on the post-retrieval selection mechanism, entropy-based uncertainty sampling for SEALS and AnchorAL and most-likely-positive selection for the active-search SEALS variant, against \system{}'s adaptive RankNet/QBC ranking.  GAL and PF-MA, the two most recent single-class discovery methods, are the closest in problem setting, so the comparison tests \system{} against the current state of that setting.    The remaining methods above (BADGE, TypiClust, GALAXY, ProbCover, DropQuery) are not included as baselines.  The deciding factor is how much adaptation our protocol would force on a method.  SEALS and AnchorAL natively address rare-class binary problems with the same pre-filter-then-select structure, and GAL and PF-MA are built for single-class discovery, so SEALS, AnchorAL, GAL and PF-MA keep their published selection rules, adapted to our shared cold start (Section~\ref{sec:experiments}).  Each excluded method instead differs along several axes simultaneously (per-round deep retraining and gradient structure for BADGE, typicality- and coverage-based selection paradigms for TypiClust and ProbCover, dropout-ensemble selection for DropQuery, and per-class graph construction with bisection selection designed around per-round deep retraining for GALAXY), so adapting one would require design decisions that could unfairly advantage or disadvantage either side, and performance differences could not be attributed to any single choice.
\section{Method}\label{sec:method}

Starting from a single positive and a single negative seed for a target class~$c$ (the positive seed may come from an external source, and the negative seed can be any random sample), \system{} iterates through three stages: retrieval, ranking, and feedback.  Figure~\ref{fig:pipeline} illustrates the architecture and Algorithm~\ref{alg:funnel} gives the complete procedure.

\begin{figure*}[t]
  \centering
  \resizebox{0.85\textwidth}{!}{%
  \begin{tikzpicture}[
    >=Stealth,
    node distance=1.6cm and 1.8cm,
    every node/.style={font=\small},
    box/.style={
      draw, rounded corners=4pt, minimum height=1.1cm,
      minimum width=2.4cm, align=center, text width=2.2cm,
      line width=0.7pt
    },
    phase/.style={
      draw, rounded corners=3pt, minimum height=1.4cm,
      minimum width=2.6cm, align=center, text width=2.5cm,
      line width=0.6pt, font=\footnotesize
    },
    arrow/.style={->, line width=0.8pt},                            %
    carrow/.style={->, line width=0.8pt, dashed},                   %
    sarrow/.style={->, line width=0.8pt, densely dotted},           %
    detail/.style={-, line width=0.5pt, black!30, dashed},          %
    label/.style={font=\scriptsize, text=black!60, align=center},
  ]

  \node[box, fill=purple!15, text width=2.0cm, minimum width=2.2cm] (seed)
    {\textbf{Cold Start}};

  \node[box, fill=blue!18, right=of seed] (retrieval)
    {\textbf{Stage 1}\\[1pt]Retrieval\\(DWVA-KNN)};

  \node[box, fill=red!15, right=of retrieval, text width=2.6cm, minimum width=2.8cm] (ranking)
    {\textbf{Stage 2}\\[1pt]Adaptive\\Ranking};

  \node[box, fill=cyan!15, right=of ranking] (oracle)
    {\textbf{Oracle}\\[1pt]Annotation\\$k$ labels/iter};

  \node[box, fill=blue!35, text=white, right=2.4cm of oracle, text width=2.0cm, minimum width=2.2cm] (classifier)
    {\textbf{Output}\\[1pt]$\mathcal{P}, \mathcal{N}$\\+ classifier $h$};

  \draw[arrow] (seed) -- (retrieval)
    node[midway, above, label] {$1{+}1$ seeds};
  \draw[arrow] (retrieval) -- (ranking)
    node[midway, above, label] {$\mathcal{U}\!\to\!\mathcal{C}$\\(top 5k)};
  \draw[arrow] (ranking) -- (oracle)
    node[midway, above, label] {top-$k$ batch};
  \draw[carrow] (oracle) -- (classifier)
    node[midway, above, label] {at the\\budget};

  \draw[sarrow] (oracle.south) -- ++(0,-0.9) -| (retrieval.south)
    node[pos=0.5, below, label] {feedback: update $\mathcal{P},\,\mathcal{N},\,\mathcal{U}$};

  \path (seed) -- (classifier) coordinate[midway] (topmid);

  \node[phase, fill=red!10, text width=3.0cm, minimum width=3.2cm] (p1)
    at ($(topmid) + (-2.0cm,-4.0cm)$)
    {\textbf{Phase 1}\\[2pt]RankNet-only\\pure exploitation\\$k \!\leftarrow\! \mathrm{round}(\gamma k)$\\after an all-positive batch};

  \node[phase, fill=orange!12, right=1.8cm of p1] (p2)
    {\textbf{Phase 2}\\[2pt]60\% RankNet\\+ 40\% QBC\\$k$ fixed};

  \draw[carrow, black!60] (p1) -- (p2)
    node[midway, above, label] {avg prec$_{w} < \tau$\\(one-way)};

  \draw[detail] (ranking.south) -- ++(0,-0.55) -- ($(p1.north)+(0,0.35)$) -- (p1.north);

  \begin{scope}[on background layer]
    \node[draw=black!20, rounded corners=6pt, fill=black!2,
          fit=(p1)(p2), inner sep=12pt] (phasebox) {};
  \end{scope}
  \node[font=\small\bfseries, text=black!70, anchor=north west]
    (phaselabel) at ([xshift=-1pt, yshift=12pt]phasebox.north west)
    {Stage 2 detail: precision-triggered phases};

  \coordinate (lg) at ($(phasebox.east)+(1.4cm,0.75cm)$);
  \draw[arrow]  (lg) -- ++(0.8,0)
    node[right, font=\scriptsize, text=black!70] {data flow};
  \draw[carrow] ($(lg)+(0,-0.5)$) -- ++(0.8,0)
    node[right, font=\scriptsize, text=black!70] {control / trigger};
  \draw[sarrow] ($(lg)+(0,-1.0)$) -- ++(0.8,0)
    node[right, font=\scriptsize, text=black!70] {state update};
  \begin{scope}[on background layer]
    \node[draw=black!20, rounded corners=3pt, fill=white, inner sep=6pt,
          fit={(lg) ($(lg)+(2.9,0.25)$) ($(lg)+(0,-1.15)$)}] (legendbox) {};
  \end{scope}

  \end{tikzpicture}%
  }%
  \caption{The \system{} pipeline.  Solid edges carry data (seeds, candidate set~$\mathcal{C}$, ranked batches).  Dashed edges carry control signals (the one-way precision trigger, termination at the budget).  Dotted edges are state updates (annotator feedback refreshing $\mathcal{P}$, $\mathcal{N}$, $\mathcal{U}$).  The lower box expands Stage~2: pure RankNet exploitation, with the batch growing after every all-positive batch, switching permanently to the 60/40 RankNet/QBC hybrid when the rolling batch precision drops below~$\tau$.}
  \Description{Flowchart of the FunnelAL pipeline showing the iterative loop through retrieval, ranking, annotation, and feedback, with a legend distinguishing data-flow, control, and state-update edges, and a detail box expanding the two precision-triggered ranking phases.}
  \label{fig:pipeline}
\end{figure*}

\subsection{Problem Setting}\label{sec:method:setting}

Let $\mathcal{X} = \{x_1, \dots, x_N\}$ be a corpus of $N$ unlabeled items with embeddings $\mathbf{x}_i \in \mathbb{R}^d$, and let $c$ be a single target class of interest.  Each item carries an unknown binary label (positive if the item belongs to~$c$), and positives form a small minority of the corpus (prevalence ranges from 0.006\% to 5.4\% in our datasets).  The annotator supplies one verified positive seed $p_0$ and one negative seed $n_0$, with no taxonomy of other classes and no cross-class supervision available.  At each round, the system recommends a batch of $k$ items from the unlabeled pool~$\mathcal{U}$, the annotator labels each one, and the labeled sets $\mathcal{P}$ and $\mathcal{N}$ grow accordingly, until an annotation budget $B$ is exhausted (in our simulation, also once every positive has been queried, Section~\ref{sec:experiments}).  The goal is twofold: (i)~\emph{discovery}, labeling as many of the corpus's true positives as possible, and (ii)~\emph{generalization}, producing a labeled set from which a classifier attains high F1 on unseen data, both with as few labels and annotation rounds as possible.  In recommendation terms, the corpus is the catalog, the annotator is the user whose ``preference'' is membership in~$c$, each round is a session, and the system's task is to maximize what each interaction teaches it.

\subsection{Stage 1: Retrieval}\label{sec:method:retrieval}

Given a corpus of $N$ images represented by $\ell_2$-normalized embeddings $\{\mathbf{x}_i\}_{i=1}^{N} \subset \mathbb{R}^d$, the retrieval stage narrows the unlabeled pool~$\mathcal{U}$ to a manageable candidate set~$\mathcal{C}$ using the current positive set~$\mathcal{P}$ as queries.

We build a FAISS~\citep{johnson2021faiss} inner-product index over all working-set embeddings.  For each positive $\mathbf{x}_p \in \mathcal{P}$, the retrieval stage retrieves its $K$ nearest neighbors in the working set and keeps those in~$\mathcal{U}$.  A query therefore contributes at most $K$ candidates (at most 511 at the first round, since a query finds itself).  Retrieving from the unlabeled pool alone changes no value of \system{} at the budget by more than 0.003 (Appendix~\ref{app:unlabeled-pool}).  Each retrieved candidate~$j$ is scored by a Distance-Weighted Vote Aggregation (DWVA) measure:
\begin{equation}\label{eq:dwva}
  s_j \;=\; \frac{|\{p \in \mathcal{P} : j \in \mathrm{kNN}(p)\}|}{|\mathcal{P}|}
  \;\cdot\;
  \left(\frac{1}{1 + \bar{d}_j}\right)^{\!\alpha},
\end{equation}
where $\bar{d}_j$ is the mean cosine distance between~$j$ and all positives that retrieved it, and $\alpha=2$ controls the distance weighting.

Candidates are sorted by~$s_j$ and truncated to the top~$C$ (we use $K{=}512$ neighbors per query and $C{=}5{,}000$).  $K$ and $C$ are tunable parameters.  The values used here were chosen for the scale of our benchmark corpora and should be adjusted for substantially different corpus sizes.  For consistency, we fix $K$ and $C$ across all five datasets, even though $C$ exceeds the pool size of UC Merced.  On UC Merced, rerunning \system{} with $K{=}256$ and $C{=}1{,}000$ changes no value at the budget by more than 0.006 and leaves its position unchanged.  The one consistent change is a slightly higher AULC-PosR (0.805 to 0.811, 20 of 21 classes).  With $K{=}256$ the candidate set never reaches 1,000, so the cap $C$ never applies and this rerun in effect changes only $K$.  On iNaturalist we also ran a wider retrieval, $K{=}2{,}000$ and $C{=}20{,}000$ (Appendix~\ref{app:k2000}).

The retrieval stage plays the role of \emph{candidate generation} in industrial recommender systems~\citep{covington2016deep}.  It costs one batched FAISS search per round.  On a corpus larger than~$C$ it also bounds the ranking stage, which then scores at most $C$ candidates instead of the whole unlabeled pool.

Retrieval comes back empty when every neighbor of the labeled positives is already labeled, and the batch is then drawn at random from the pool.  In our runs this happens only on the large corpora, in tasks that never find a second positive.  The warm-up labels the seed's 511 neighbors within 26 rounds, and the remaining rounds are random (3 of 100 iNaturalist tasks with DINOv2, Section~\ref{sec:limitations}).

\subsection{Stage 2: Ranking}\label{sec:method:ranking}

Within the candidate set~$\mathcal{C}$, the system applies two ranking strategies, corresponding to the exploitation and exploration arms.

\paragraph{Exploitation: RankNet.}
A three-layer neural network $f_\theta : \mathbb{R}^d \!\to\! \mathbb{R}$ ($d$--128--128--1 with ReLU activations, where $d$ is the embedding dimensionality) is trained with the pairwise ranking loss of \citet{burges2005learning}:
\begin{equation}\label{eq:ranknet}
  \mathcal{L}
  = \sum_{(p,n) \in \mathcal{P}\times\mathcal{N}}
    \log\!\bigl(1 + e^{-(f_\theta(\mathbf{x}_p) - f_\theta(\mathbf{x}_n))}\bigr),
\end{equation}
where $\mathcal{P}$ and $\mathcal{N}$ are the current positive and negative sets.    The model is retrained from scratch every round, for 30 epochs with Adam~\citep{kingma2015adam} ($\mathrm{lr}{=}10^{-3}$).  Training requires at least two labeled positives and two labeled negatives.  Each epoch draws $n$ pairs uniformly with replacement from $\mathcal{P}\times\mathcal{N}$, with $n = \min(4096, \max(256, |\mathcal{P}|\cdot|\mathcal{N}|))$, and takes one gradient step on their mean loss.  Candidates are
ranked by $\sigma(f_\theta(\mathbf{x}_j))$, where $\sigma$ is the logistic sigmoid, and the top-$k$ are recommended. Directly analogous to \emph{exploitation} in bandit
recommendation~\citep{li2010contextual}, this strategy surfaces items most likely to be positive.

\paragraph{Exploration: Query-by-Committee.}
A two-member linear committee is trained on $\mathcal{P} \cup \mathcal{N}$, with features standardized on the labeled set: logistic regression ($C{=}1.0$, balanced class weights) and a linear SVM ($C{=}1.0$, balanced class weights, decision values mapped through a sigmoid).  Each candidate~$j$ receives a committee uncertainty score $h_j$ defined as the binary entropy of the committee's mean prediction:
\begin{equation}\label{eq:qbc}
  h_j = -\bigl[\bar{p}_j \log \bar{p}_j + (1{-}\bar{p}_j)\log(1{-}\bar{p}_j)\bigr],
\end{equation}
where $\bar{p}_j$ is the mean positive-class probability across committee members.  This score measures uncertainty in the committee's averaged prediction rather than disagreement among its members.  Candidates with the highest scores have averaged predictions closest to $0.5$ and are prioritized for annotation.  We retain the label QBC for this committee-based component, specifying that its criterion is predictive entropy rather than a disagreement measure.  This strategy parallels \emph{exploration} in bandit systems~\citep{chapelle2011empirical}.

RankNet starts training with very few labels, and pairs built from so few examples could mislead it.  Three properties of the design limit this risk.  RankNet only reranks the retrieved candidate set, which is already enriched in positives.  The ranker is retrained from scratch each round, so no early fit persists.  And the precision trigger reads the batches an early ranker selects, so a ranker that performs poorly moves the system to the mixed phase.

\subsection{Hybrid Strategy}\label{sec:method:adaptive}

In \system{}, the annotator may freely switch between exploitation and exploration at each iteration based on their observation of the recommended samples. For instance, switching to exploration when the suggested positives become less convincing, or inspecting the top-$k$ lists from both RankNet and QBC side by side to accelerate annotation.  Because this interactive mode depends on subjective annotator judgment that cannot be faithfully simulated with an oracle, we additionally propose a \emph{hybrid} policy that automates the choice via a precision-triggered transition, enabling reproducible evaluation.  Once sufficient positives have been collected (see Algorithm~\ref{alg:funnel} for warm-up details), the policy proceeds through two phases:

\begin{enumerate}
    \item \textbf{Phase 1: Pure exploitation.}  The system uses RankNet exclusively.  After every all-positive batch, the batch size grows by a factor $\gamma{=}1.2$, rounded to the nearest integer ($k \leftarrow \mathrm{round}(\gamma \cdot k)$).  Growth stops at the switch.  The precision of each batch selected in this phase is recorded, and the system moves to Phase~2 when the mean over the last $w{=}3$ recorded batches falls below a transition threshold $\tau{=}0.7$.  Warm-up batches are not recorded.  If the warm-up batch is all positive, RankNet cannot yet train (fewer than two labeled negatives), and the next batch is selected without a trained ranker.  Its scores are then all equal, and the batch is the first $k$ entries of \texttt{np.argsort(scores)[::-1]}, with the candidates in descending DWVA order and the scores as float64 zeros.  NumPy's default sort is not stable and does not define the order of equal values.  In the recorded experimental environment, with NumPy 1.24.4, 512 candidates and $k{=}20$ produced retrieval ranks 162 to 179, 511 and 512, mostly candidates about a third of the way down the list.  Its precision is recorded like the others.  Appendix~\ref{app:trigger-window} reports a variant that skips this batch and counts the warm-up batch instead.
    \item \textbf{Phase 2: Hybrid exploit/explore.}  Each batch of $k$ recommendations is split into 60\% from RankNet and 40\% from QBC, applied to disjoint subsets of~$\mathcal{C}$.
\end{enumerate}
The values of $\tau$, $w$, and $\gamma$ were not tuned per dataset.  The same defaults are used across all experiments.

\subsection{Final Classifier}\label{sec:method:classifier}

The primary output of \system{} is the labeled sets $\mathcal{P}$ and $\mathcal{N}$, which can be used to train any downstream model suited to the application.  For evaluation, we train a logistic regression classifier on the accumulated labels at every round.  Because the funnel preferentially discovers positives, the labeled set can be highly imbalanced.  We mitigate this by augmenting the training set with a random sample of unlabeled items treated as presumed negatives ($\max(2{,}000,\, 5\times$ labeled positives$)$ of them, or the whole unlabeled pool when it is smaller), following a positive-unlabeled learning approach~\citep{bekker2020pu}, and using balanced class weights.  Evaluation details are described in Section~\ref{sec:experiments}.

\begin{algorithm}[t]
\caption{\system{}: Adaptive Funnel for Active Annotation}
\label{alg:funnel}
\footnotesize
\begin{algorithmic}[1]
\Require Corpus $\mathcal{X}$, embeddings $\{\mathbf{x}_i\}$, class $c$, budget $B$, initial batch $k_0$
\Require Transition threshold $\tau$ (default $0.7$), window size $w$ (default $3$), growth factor $\gamma$ (default $1.2$)
\Ensure Labeled sets $\mathcal{P}, \mathcal{N}$; classifier $h$
\Statex \textit{$\mathrm{top}_m(\mathcal{A})$: the $m$ highest-scoring items of $\mathcal{A}$, or all of $\mathcal{A}$ if it holds fewer.  When $\mathcal{C} \neq \emptyset$, every batch comes from $\mathcal{C}$ alone and holds $\min(k, |\mathcal{C}|)$ items.}
\State \textbf{Cold start:} annotator provides one positive seed $p_0$ and one negative seed $n_0$
\State $\mathcal{P} \leftarrow \{p_0\},\;\; \mathcal{N} \leftarrow \{n_0\},\;\; \mathcal{U} \leftarrow \mathcal{X} \setminus \{p_0, n_0\}$
\State $k \leftarrow k_0$;\quad $\mathrm{hybrid\_triggered} \leftarrow \mathrm{False}$;\quad $\mathrm{prec\_history} \leftarrow [\,]$
\While{$|\mathcal{P}|{+}|\mathcal{N}| < B + 2$ \textbf{and} $\mathcal{U} \neq \emptyset$} \Comment{the two seeds are not counted against $B$}
  \State $k \leftarrow \min(k, |\mathcal{U}|)$
  \State \textit{// Stage 1: Retrieval (DWVA)}
  \State $\mathcal{C} \leftarrow \mathrm{DWVA\text{-}KNN}(\mathcal{U}, \mathcal{P})$ \Comment{Eq.~\ref{eq:dwva}}
  \State \textit{// Stage 2: Precision-triggered adaptive ranking}
  \If{$\mathcal{C} = \emptyset$} \Comment{random fallback (Section~\ref{sec:method:retrieval})}
    \State $\mathcal{S} \leftarrow$ $k$ items drawn uniformly without replacement from $\mathcal{U}$
  \ElsIf{$|\mathcal{P}| < 2$} \Comment{warm-up}
    \State $\mathcal{S} \leftarrow \mathrm{DWVA\text{-}top}_k(\mathcal{C})$
  \Else
    \If{$\neg\,\mathrm{hybrid\_triggered}$}
      \If{the preceding batch was selected in Phase 1}
        \State Append its precision to $\mathrm{prec\_history}$
      \EndIf
      \If{$|\mathrm{prec\_history}| \geq w$ \textbf{and} $\mathrm{mean}(\mathrm{prec\_history}[-w:]) < \tau$}
        \State $\mathrm{hybrid\_triggered} \leftarrow \mathrm{True}$
      \EndIf
    \EndIf
    \If{$\neg\,\mathrm{hybrid\_triggered}$} \Comment{Phase 1: exploitation}
      \State $\mathcal{S} \leftarrow \mathrm{RankNet\text{-}top}_k(\mathcal{C}, \mathcal{P}, \mathcal{N})$ \Comment{while $|\mathcal{N}| < 2$ RankNet is untrained and all scores tie (Section~\ref{sec:method:adaptive})}
    \Else \Comment{Phase 2: exploit/explore}
      \State $m \leftarrow \max(1, \lfloor 0.6k \rfloor)$
      \State $\mathcal{S}_{\mathrm{E}} \leftarrow \mathrm{RankNet\text{-}top}_{m}(\mathcal{C}, \mathcal{P}, \mathcal{N})$
      \State $\mathcal{S}_{\mathrm{X}} \leftarrow \mathrm{QBC\text{-}top}_{k - m}(\mathcal{C} \!\setminus\! \mathcal{S}_{\mathrm{E}}, \mathcal{P}, \mathcal{N})$
      \State $\mathcal{S} \leftarrow \mathcal{S}_{\mathrm{E}} \cup \mathcal{S}_{\mathrm{X}}$
    \EndIf
  \EndIf
  \State \textit{// Stage 3: Feedback}
  \For{each $x \in \mathcal{S}$}
    \If{annotator labels $x$ positive}
      \State $\mathcal{P} \leftarrow \mathcal{P} \cup \{x\}$
    \Else
      \State $\mathcal{N} \leftarrow \mathcal{N} \cup \{x\}$
    \EndIf
    \State $\mathcal{U} \leftarrow \mathcal{U} \setminus \{x\}$
  \EndFor
  \If{$\neg\,\mathrm{hybrid\_triggered}$ \textbf{and} every $x \in \mathcal{S}$ was labeled positive}
    \State $k \leftarrow \mathrm{round}(\gamma \cdot k)$ \Comment{grow the next batch}
  \EndIf
\EndWhile
\State Train classifier $h$ on $\mathcal{P}$ (pos) $\cup$ $\mathcal{N}$ $\cup$ $\mathrm{sample}(\mathcal{U})$ (neg)
\Return $\mathcal{P}, \mathcal{N}, h$
\Statex \textit{In the experiments a run also ends once every positive has been queried, an oracle rule (Section~\ref{sec:protocol}).  The last batch is not cut at the budget, so a run can end with more than $B$ labels after the seeds.  Every value is read at the budget, $B$ labels after the two seeds.}
\end{algorithmic}
\end{algorithm}

\section{Experiments}\label{sec:experiments}

\subsection{Datasets and Embeddings}\label{sec:datasets}

We evaluate on five datasets.  Three benchmarks vary in \emph{task granularity} and in \emph{domain distance} from the encoder's pre-training distribution.  Two large corpora test the setting the funnel is designed for: a rare class in a corpus of up to 2.7 million images.

\paragraph{CUB-200-2011~\citep{wah2011caltech}.}
A fine-grained bird species dataset containing 11,788 variable-resolution images across 200 species (${\sim}60$ images per class).  While most species are reasonably separable in embedding space, confusability is concentrated among closely related groups (e.g., gulls, terns) that share plumage patterns and demand fine-grained discrimination. Class prevalence is 0.5\%.

\paragraph{FGVC-Aircraft~\citep{maji2013fgvc}.}
A fine-grained benchmark of variable-resolution aircraft images.  The official release (fgvc-aircraft-2013b) contains 10,000 images of 100 model variants.  Our label preprocessing pooled variants whose names share their leading word, yielding 90 classes: 83 single model variants, plus 7 coarser groups that absorb the remaining 17 variants (e.g., the four Cessna variants become one class, while single-token names such as 737-800 were unaffected).  We retain this labeling so that every experiment in the paper uses the same classes.  Most classes have ${\sim}100$ images and the pooled ones 200--400.  Variants are distinguished by subtle structural cues such as winglets and engine nacelles. Mean class prevalence is 1.1\%.

\paragraph{UC Merced Land Use~\citep{yang2010ucmerced}.}
An aerial imagery benchmark of 2,100 RGB images ($256{\times}256$ pixels at ${\sim}0.3$\,m ground resolution) from the USGS National Map, organized into 21 land-use classes with 100 images each.  The overhead imaging modality represents a domain shift from the web-crawled photographs that dominate pre-training corpora, though the encoder's learned visual primitives still transfer and produce informative embeddings. Class prevalence is 4.8\%.  

\paragraph{iNaturalist 2021~\citep{vanhorn2021inat}.}
The full training split, 2,686,843 photographs of 10,000 species.  We evaluate 20 target species, chosen before any run and stratified by class size and taxon, with 157 to 300 images each.  Prevalence is 0.006\% to 0.011\%.  The difficulty is rarity: one seed positive against a working set of 2.15 million images.

\begin{figure}[t]
  \centering
  \includegraphics[width=\columnwidth]{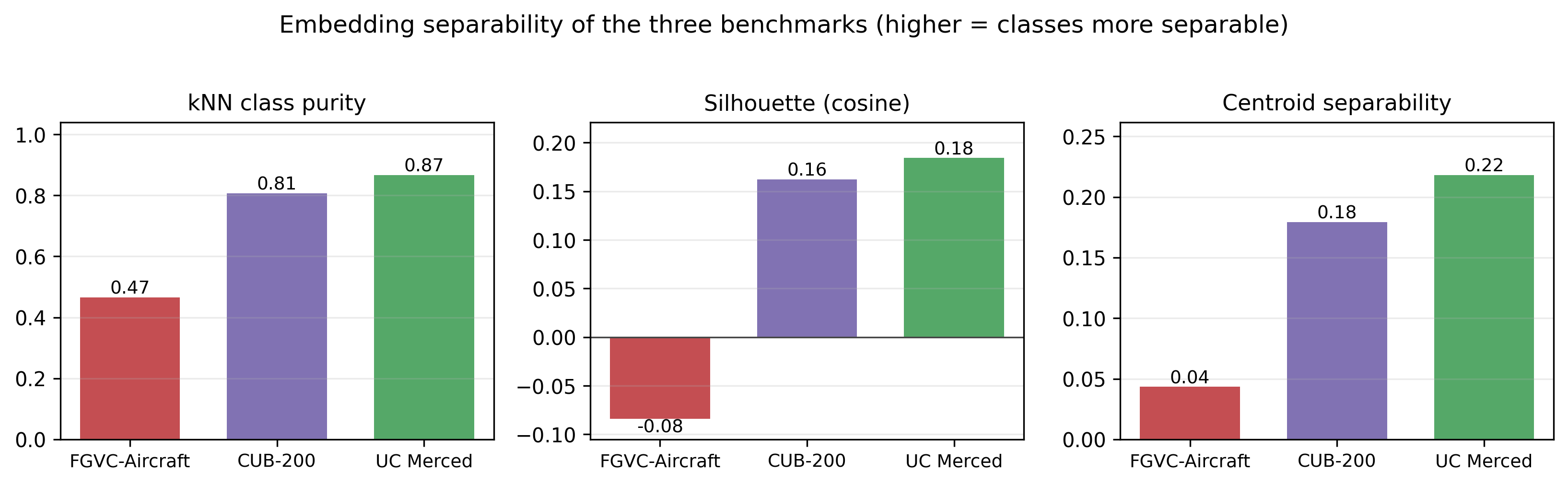}
  \caption{Embedding separability of the three benchmarks, computed from the DINOv2 embeddings and labels alone (no annotation method involved).  \emph{Left:} kNN class purity, the fraction of each image's 10 nearest neighbors that share its class.  \emph{Center:} silhouette coefficient (cosine)~\citep{rousseeuw1987silhouettes}.  \emph{Right:} centroid separability, one minus the mean cosine similarity between each class centroid and its nearest other-class centroid.  All three measures agree.  UC~Merced's land-use classes are the most separable and FGVC-Aircraft's variants the least.  FGVC's negative silhouette indicates that the average aircraft image lies closer to a different variant's cluster than to its own.  Because the retrieval stage relies on neighbors sharing the class of interest, it breaks down on the least separable benchmark, FGVC-Aircraft, where the ranking stage becomes essential.}
  \Description{Three bar charts, one per separability measure, each showing FGVC-Aircraft lowest, CUB-200 middle, and UC Merced highest.}
  \label{fig:separability}
\end{figure}

\paragraph{fMoW~\citep{christie2018fmow}.}
The training split of fMoW-rgb: 83,412 functional sites in 62 categories, one image per site (the first of its views).  We evaluate 10 target categories, chosen before any run to span a wide range of prevalence (44 to 4,521 images, 0.05\% to 5.4\%).  Overhead imagery lies outside the encoder's pre-training domain, and its classes are less separable in the embedding than those of iNaturalist.  Their kNN class purity, the measure of Figure~\ref{fig:separability} computed on each corpus, is 0.369 against 0.556.  Neither corpus appears in the published composition of DINOv2's pre-training set (Table~15 of \citet{oquab2024dinov2}).  That table does not list every retrieved image, so this is absence from the published composition, not a verified absence.

\medskip
All images are encoded using DINOv2~ViT-S/14 with registers~\citep{oquab2024dinov2, darcet2024registers}, producing 384-dimensional embeddings.  The retrieval stage uses $\ell_2$-normalized copies of them (Section~\ref{sec:method:retrieval}), and RankNet uses the raw vectors.  Each corpus is split 80/20 via stratified sampling into a working set and a held-out test set.  The test set is never used during annotation.  The working sets of iNaturalist and fMoW hold 2,149,474 and 66,729 images.  Appendix~\ref{app:clip} repeats the main comparison with a second encoder, CLIP ViT-B/16~\citep{radford2021clip}, with 512-dimensional embeddings.  It covers all ten methods on the five datasets, with clean labels and at 10\% noise.  The two encoders were trained differently: DINOv2 by self-supervision on images, CLIP on paired images and text.

\begin{figure*}[t]
  \centering
  \includegraphics[width=0.8\textwidth]{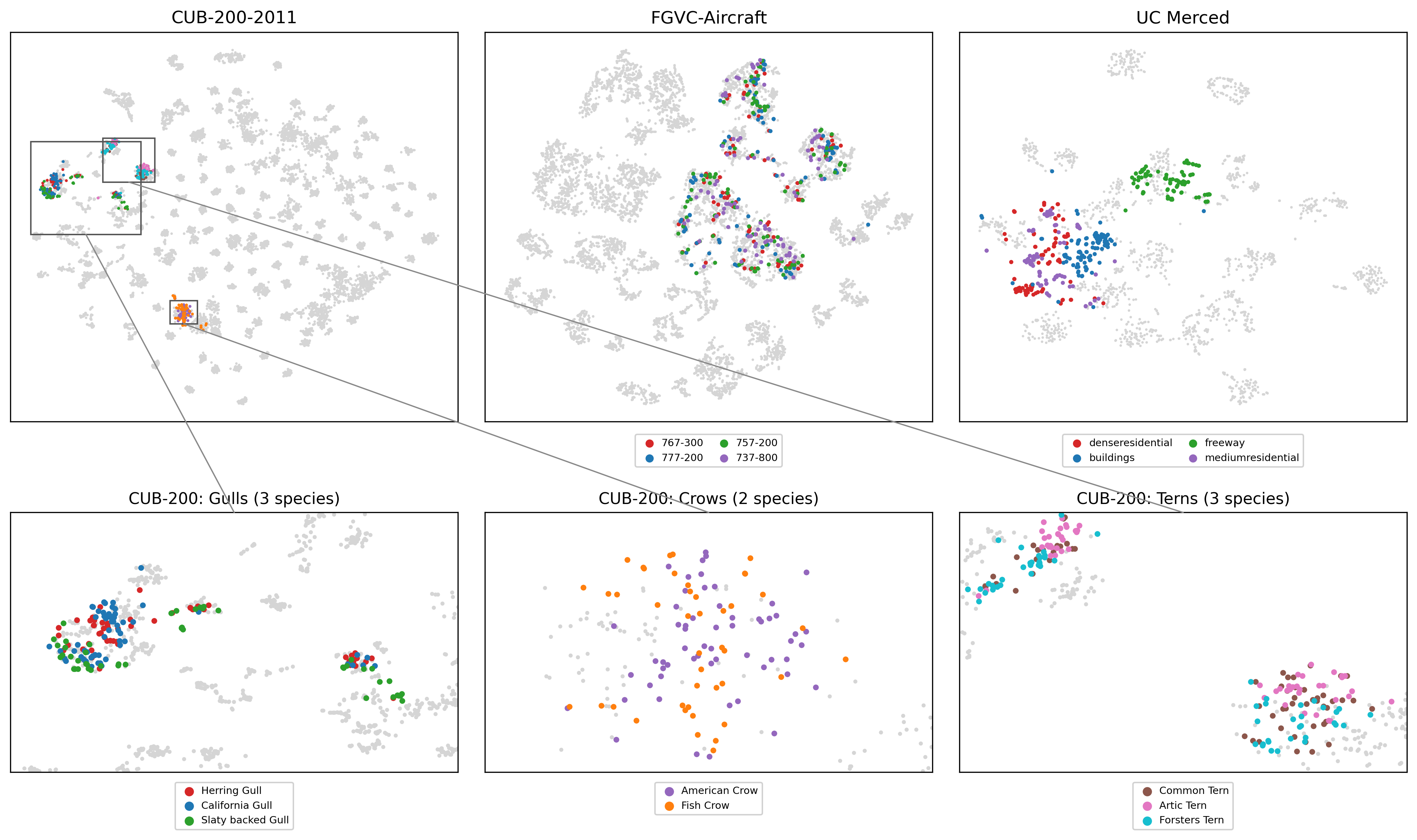}
  \caption{t-SNE~\citep{vandermaaten2008tsne} of the DINOv2 embeddings, illustrating the differing class structure of the three benchmarks (computed from the embeddings alone, with no annotation method involved).  \emph{Top row:} each dataset with some representative confusable classes highlighted (three gull, two crow and three tern species for CUB-200, four Boeing variants for FGVC-Aircraft, and the four least-separable land-use classes for UC~Merced, ranked by the kNN class purity of Figure~\ref{fig:separability}), with all other samples in light gray.  \emph{Bottom row:} zoomed views of the three CUB-200 confusable genera, showing that the embedding groups them by genus but does not separate the species within.  Aircraft variants form heavily overlapping clusters and bird species overlap within confusable genera, whereas even UC~Merced's least-separable classes remain far more coherent, with overlap confined to the residential-density boundary, illustrating why similarity-based retrieval alone struggles on the fine-grained benchmarks.  As with any t-SNE view, only local structure is faithful.  Whether nearby points mingle is meaningful, while distances between far-apart clusters are not.}
  \Description{Three t-SNE scatter plots. CUB-200 and FGVC-Aircraft show highlighted classes forming overlapping clusters.  UC Merced shows its highlighted classes as more coherent regions whose overlap is confined to the residential classes.}
  \label{fig:tsne}
\end{figure*}

The three benchmarks span a wide range of embedding separability, visible qualitatively in the embedding structure (Figure~\ref{fig:tsne}) and quantified by three intrinsic measures (Figure~\ref{fig:separability}).  In embedding space, aircraft variants form heavily overlapping clusters, and bird species overlap mainly within confusable genera.  The embedding separates coarse groups (genus, manufacturer), but not always the fine categories within them.  UC~Merced's land-use classes, by contrast, separate into largely coherent islands.  All three measures agree on the ordering.  UC~Merced's classes are the most cleanly separated and FGVC-Aircraft's variants the least, with CUB-200 in between.  FGVC's negative silhouette reflects variants so entangled that the average image is closer to a wrong variant's cluster than to its own.  Since the retrieval stage depends on nearest neighbors sharing the target class, these gaps anticipate its behavior.  Retrieval is far more effective on UC~Merced than on FGVC-Aircraft, where entangled variants cause similarity search to break down and the ranking stage must carry the load.  Even on UC~Merced, retrieval alone does not separate the most confusable classes (Retrieval-only finds 84\% of the positives, Table~\ref{tab:ablations}).  High separability instead means the exploit-capable methods cluster near the ceiling, making it the setting where margins among the leaders are smallest.

\system{} targets single-class discovery: one class of interest must be found in a large corpus starting from minimal supervision.  The one-vs-rest evaluation protocol simulates this setting at class prevalences ranging from 0.006\% to 5.4\%.  The five datasets span a range of domain distance from the encoder's pre-training distribution, and \system{}'s retrieve-then-rank architecture is evaluated across this entire spectrum.

\subsection{Baselines and Ablations}

We compare \system{} with nine strategies.  Three are \emph{ablations} that remove one component of \system{}, and six are \emph{external baselines}.  GAL is compared on the three benchmarks in Appendix~\ref{app:gal}.  Its selection takes 34\,s per round on FGVC-Aircraft, against 0.14\,s for \system{}, which rules out the large corpora.

\paragraph{Ablations.}
\begin{itemize}
    \item \textbf{Retrieval-only:} Stage~1 only (DWVA-scored KNN, Eq.~\ref{eq:dwva}) with no Stage~2 ranking.  Tests whether retrieval alone suffices.
    \item \textbf{Funnel-RankNet:} DWVA retrieval + RankNet ranking (pure exploitation, no QBC).
    \item \textbf{Funnel-QBC:} retrieval + QBC ranking (pure exploration, no RankNet).
\end{itemize}
\noindent Each ablation uses \system{}'s warm-up and batch growth.  With no switch, its batch keeps growing after every all-positive batch.

\paragraph{External baselines.}
\begin{itemize}
    \item \textbf{Random:} $k$ samples drawn uniformly from the pool.  The standard lower-bound baseline in active learning.
    \item \textbf{Uncertainty:} one logistic regression, fitted each round on features standardized with the labeled set, selects the items whose predicted probability is closest to 0.5, over the whole pool.  A standard active learning baseline~\citep{settles2012active}.
    \item \textbf{SEALS}~\citep{coleman2022similarity}: candidates are the $k{=}100$ nearest unlabeled neighbors of the labeled set, found by exact cosine search.  The neighbors of each pick join the candidates before the next pick (Algorithm~2 of the paper).  A logistic regression on $\ell_2$-normalized embeddings, fitted once per round, selects by maximum entropy.  The paper uses approximate search and batches of 100.
    \item \textbf{SEALS (most-likely-pos.)}~\citep{coleman2022similarity}: the same candidates, selecting the highest predicted positive probability, the paper's active-search mode.
    \item \textbf{AnchorAL}~\citep{lesci-vlachos-2024-anchoral}: each round, 10 anchors per class are drawn from the labeled set by k-means++, or all labeled items of a class while it has 10 or fewer.  From our one-positive start, the first round therefore has one anchor per class.  The 50 nearest unlabeled neighbors of each anchor form a subpool, ranked by mean cosine similarity.  A logistic regression on $\ell_2$-normalized embeddings selects from it by maximum entropy.  The paper's learner is a fine-tuned transformer.  We build the subpool every round, as in its Algorithm~1.  The official code builds it only for pools of more than 10,000 items, which on our benchmarks would make AnchorAL nearly identical to Uncertainty (the same entropy ranking over the whole pool, on differently scaled features).
    \item \textbf{PF-MA}~\citep{zaher2026pfma}: the positive-first most-ambiguous score (Eq.~3 of the paper) over every unlabeled item, from a linear SVM refitted each round on $\ell_2$-normalized embeddings.  All predicted positives rank above all predicted negatives, most-ambiguous positives first.  Decision values are mapped to $[0,1]$ by a sigmoid.  The paper does not state the mapping.
\end{itemize}

\noindent Every method starts from the same seed pair, one positive and one negative, and labels 20 items per round.  For the baselines this is an adaptation.  Their papers start from larger labeled sets: 5 positives and 95 negatives for SEALS, 100 items including 5 from each minority class for AnchorAL, and 1 positive and 5 negatives for PF-MA.  Each baseline keeps its published selection rule and the settings of its paper, apart from the shared start, the batch size and the departures listed above.  We did not tune any method's numeric settings on our data.  Where a paper leaves a design choice open, we ran the alternatives and report the one that did better for that baseline.  Where a paper fixes a choice, we follow it.  We checked each baseline against its source.  GAL's code offers two ways to score a candidate.  The default keeps only the first two coordinates of the classifier's weight vector, which suits two-dimensional toy data.  It also has a copy-paste error that computes the change under a positive pseudo-label from the fit under a negative one.  We follow the other option, \texttt{slope\_only}, which compares full weight vectors, and on the same inputs our GAL picks the same items as that option.  AnchorAL's anchor sampling and subpool search match the authors' code on the same inputs.  We found no code released by the authors of SEALS or PF-MA, so we checked our implementations against the algorithms and equations of their papers.

\subsection{Evaluation Protocol}\label{sec:protocol}

On the three benchmarks each strategy is evaluated on every class.  On iNaturalist and fMoW it is evaluated on the target classes of Section~\ref{sec:datasets}.  Each class is run 5 times with different random seeds, which change the seed pair and the strategies' random choices.  Every strategy labels 20 items per round.  \system{} and its ablations grow the batch as described in Section~\ref{sec:method:adaptive}.  On the benchmarks the budget is three times the number of positives in the working set, $B = 3\,n_{\mathrm{pos}}$.  On the large corpora it is fixed: 600 labels on iNaturalist and on seven fMoW categories, and 2,000 labels on the three fMoW categories with more than 1,000 positives in the working set.  A fixed budget gives every class of a corpus the same effort and does not presume the class size, which an annotator looking for a rare class does not know.  The benchmark budget uses $n_{\mathrm{pos}}$, an oracle quantity, only to define a consistent protocol.

We simulate a \emph{perfect oracle}: each recommended sample is labeled according to its ground-truth class membership, following standard active-learning evaluation practice~\citep{settles2012active} (Section~\ref{sec:noise} relaxes this assumption with an error-prone annotator).  A run also ends when every positive in the working set has been queried.  This rule reads the ground truth, which a deployed system cannot do (Section~\ref{sec:limitations}).  Because runs end at different label counts and the last batch can pass the budget, every value is read at the budget.  A run that ended earlier keeps its last value.  A value between two rounds is interpolated linearly.

At every round, the same evaluation classifier is trained for every method: a logistic regression with balanced class weights, on features standardized with its training set.  Its training set is the labeled items plus a random sample of unlabeled items treated as negatives, $\max(2{,}000,\, 5 \times$ labeled positives$)$ of them, or the whole unlabeled pool when it is smaller.  It is scored on the held-out test set.  F1 therefore evaluates this whole training procedure, presumed negatives included.  It does not show that one method's labels are better for every downstream learner.  Unlabeled items can include positives a method has not found, and those enter as negatives.  On UC~Merced, the only dataset whose pool is smaller than 2,000 images, the whole pool is taken, so every unfound positive enters as a negative.  Re-evaluated with the labeled items alone, \system{}'s F1 on UC~Merced is unchanged (0.951 with labeled items alone, 0.951 in Table~\ref{tab:results}) and so is its mean-rank position.  The F1 gaps to methods that leave positives unfound close: SEALS 0.708 to 0.950, AnchorAL 0.724 to 0.944, Uncertainty 0.765 to 0.953 (Appendix~\ref{app:evalclf}).

We report the following metrics:

\begin{itemize}
    \item \textbf{F1} and \textbf{Recall} (test set): of the evaluation classifier at the budget.
    \item \textbf{AULC-PosR} and \textbf{AULC-F1}: the area under positive recall, or test F1, against labels, from the first round to the budget, divided by that range so that values lie in $[0,1]$.  AULC-PosR measures discovery speed, and AULC-F1 measures classifier quality along the way.
    \item \textbf{Pos.\ Rec.}\ (pool): the fraction of the working set's positives labeled by the budget.  Under noise it counts only true positives labeled positive.
    \item \textbf{Labels}: the labels used up to the budget, the two seeds included.  A method that finds every positive early uses fewer.
    \item \textbf{Mean rank}: one summary of the two goals, discovery and the classifier, with one column for each.  On each dataset the ten methods are ranked by AULC-PosR (how fast positives are found) and by F1 (the classifier at the budget), with ties sharing the average rank.  Each rank is averaged over the five datasets, and the overall mean rank is the mean of the two.  The other columns overlap with these two: Recall is part of F1, Pos.~Rec.\ is the end point of the curve under AULC-PosR, and AULC-F1 is the area under the F1 curve.
\end{itemize}

\noindent We call one run of a method on one class with one repeat a \emph{task}.  All metrics are averaged over the 5 repeats of each class, then macro-averaged across all classes within each dataset.  Run-to-run variability is small relative to the gaps between leading and trailing methods, and smaller gaps are judged by the paired tests.  The within-class standard deviation of \system{}'s final F1 across the 5 repeats averages 0.002 on UC~Merced, 0.029 on CUB-200, 0.045 on FGVC-Aircraft and 0.062 on iNaturalist (other methods range up to 0.15).

\emph{Statistical tests.}  Whenever we compare two strategies, we report a paired Wilcoxon signed-rank test~\citep{wilcoxon1945individual}, the standard test for comparing two methods across multiple tasks~\citep{demsar2006statistical}, at the class level: each one-vs-rest class contributes a single pair, its metric averaged over the 5 repeats, and one test over these per-class differences yields the reported $p$-value (alongside the win count over classes).  Pairing per class compares the two strategies on the same class, so the fact that some classes are harder than others cancels out.  Averaging the repeats before testing avoids treating them as independent samples.  For each embedding, and separately with clean labels and at 10\% noise, we test \system{} against each of the other nine methods on five columns and four datasets, 180 tests.  Testing many hypotheses raises the chance that some $p$-value falls below 0.05 by luck alone.  The standard guard, a Bonferroni correction, sets the threshold to $0.05/180 \approx 2.8{\times}10^{-4}$, and we call a difference significant only below it.  A difference that is not significant is not evidence of equivalence, and we ran no equivalence tests.  We do not test on fMoW.  It has only 10 categories, so even a method ahead in all ten would get a two-sided $p$ of $2/2^{10} \approx 0.002$, above this threshold.  For fMoW we report instead in how many of the 10 categories a method is ahead.

\subsection{Results}\label{sec:results}

Tables~\ref{tab:results} and~\ref{tab:ablations} report the values at the budget with DINOv2 and clean labels.  Table~\ref{tab:meanrank} ranks the ten methods over the five datasets.  Every comparison below is paired per class (Section~\ref{sec:experiments}), and \emph{significant} means below the corrected bar of $2.8{\times}10^{-4}$.

\begin{table}[!t]
\centering
\caption{External methods, DINOv2, clean labels.  Values at the budget (three times the class's positives on the benchmarks, 600 labels on iNaturalist), the mean over 5 repeats and then over classes.  \textbf{Bold} = best in the block and column (equal to three decimals).  Labels = labels used up to the budget, the two seeds included.  A method that finds every positive before the budget uses fewer.  fMoW's 10 categories are read at their own budgets (600 labels for seven, 2,000 for three) and averaged with equal weight, so its Labels value is a mean.}
\label{tab:results}
\footnotesize
\begin{tabular}{l cccccc}
\toprule
Strategy & F1 & Recall & AULC-PosR & AULC-F1 & Pos.\ Rec. & Labels \\
\midrule
\addlinespace
\multicolumn{7}{l}{\textit{UC Merced (21 classes)}} \\
\system{} (ours) & \textbf{0.951} & \textbf{0.963} & 0.805 & 0.841 & \textbf{0.993} & 170 \\
SEALS (most-likely-pos.) & 0.948 & 0.958 & \textbf{0.824} & \textbf{0.853} & 0.990 & 171 \\
PF-MA & 0.950 & 0.961 & 0.721 & 0.775 & 0.992 & 173 \\
SEALS & 0.708 & 0.614 & 0.307 & 0.494 & 0.514 & 242 \\
AnchorAL & 0.724 & 0.637 & 0.348 & 0.526 & 0.546 & 242 \\
Uncertainty & 0.765 & 0.702 & 0.409 & 0.594 & 0.628 & 242 \\
Random & 0.301 & 0.190 & 0.086 & 0.167 & 0.152 & 242 \\
\addlinespace
\multicolumn{7}{l}{\textit{CUB-200-2011 (200 classes)}} \\
\system{} (ours) & 0.878 & 0.889 & 0.746 & \textbf{0.823} & 0.939 & 131 \\
SEALS (most-likely-pos.) & \textbf{0.879} & \textbf{0.894} & \textbf{0.756} & 0.817 & \textbf{0.944} & 128 \\
PF-MA & 0.877 & 0.887 & 0.573 & 0.722 & 0.928 & 134 \\
SEALS & 0.778 & 0.747 & 0.204 & 0.530 & 0.380 & 143 \\
AnchorAL & 0.801 & 0.787 & 0.284 & 0.656 & 0.460 & 143 \\
Uncertainty & 0.848 & 0.852 & 0.324 & 0.654 & 0.591 & 143 \\
Random & 0.139 & 0.090 & 0.030 & 0.105 & 0.036 & 143 \\
\addlinespace
\multicolumn{7}{l}{\textit{FGVC-Aircraft (90 classes)}} \\
\system{} (ours) & \textbf{0.727} & \textbf{0.737} & \textbf{0.564} & \textbf{0.588} & \textbf{0.838} & 256 \\
SEALS (most-likely-pos.) & 0.693 & 0.678 & 0.477 & 0.532 & 0.736 & 264 \\
PF-MA & 0.716 & 0.720 & 0.452 & 0.511 & 0.800 & 263 \\
SEALS & 0.555 & 0.505 & 0.127 & 0.341 & 0.269 & 269 \\
AnchorAL & 0.555 & 0.508 & 0.205 & 0.397 & 0.376 & 269 \\
Uncertainty & 0.704 & 0.705 & 0.303 & 0.498 & 0.616 & 269 \\
Random & 0.171 & 0.110 & 0.029 & 0.102 & 0.044 & 269 \\
\addlinespace
\multicolumn{7}{l}{\textit{iNaturalist 2021 (20 species)}} \\
\system{} (ours) & 0.581 & 0.698 & \textbf{0.606} & 0.539 & 0.818 & 602 \\
SEALS (most-likely-pos.) & 0.593 & 0.708 & 0.594 & 0.493 & 0.806 & 602 \\
PF-MA & 0.715 & \textbf{0.715} & 0.525 & 0.601 & \textbf{0.839} & 602 \\
SEALS & 0.601 & 0.701 & 0.150 & 0.499 & 0.295 & 602 \\
AnchorAL & 0.451 & 0.689 & 0.202 & 0.390 & 0.356 & 602 \\
Uncertainty & \textbf{0.726} & 0.706 & 0.298 & \textbf{0.635} & 0.514 & 602 \\
Random & 0.200 & 0.220 & 0.005 & 0.208 & 0.005 & 602 \\
\addlinespace
\multicolumn{7}{l}{\textit{fMoW (10 categories)}} \\
\system{} (ours) & 0.466 & 0.418 & \textbf{0.256} & 0.368 & 0.444 & 1022 \\
SEALS (most-likely-pos.) & 0.461 & 0.416 & 0.254 & 0.366 & \textbf{0.451} & 1022 \\
PF-MA & \textbf{0.468} & 0.422 & 0.212 & 0.373 & 0.423 & 1022 \\
SEALS & 0.426 & \textbf{0.436} & 0.101 & 0.327 & 0.200 & 1022 \\
AnchorAL & 0.402 & 0.410 & 0.098 & 0.306 & 0.179 & 1022 \\
Uncertainty & \textbf{0.468} & 0.420 & 0.167 & \textbf{0.376} & 0.307 & 1022 \\
Random & 0.209 & 0.164 & 0.012 & 0.123 & 0.020 & 1022 \\
\bottomrule
\end{tabular}
\end{table}

\begin{table}[!t]
\centering
\caption{\system{} and its ablations, DINOv2, clean labels, read as Table~\ref{tab:results}.  Funnel-RankNet removes the exploration arm, Funnel-QBC the exploitation arm, and Retrieval-only the ranking stage.  $^{\dagger}$~marks ablations of \system{}.}
\label{tab:ablations}
\footnotesize
\begin{tabular}{l cccccc}
\toprule
Strategy & F1 & Recall & AULC-PosR & AULC-F1 & Pos.\ Rec. & Labels \\
\midrule
\addlinespace
\multicolumn{7}{l}{\textit{UC Merced (21 classes)}} \\
\system{} (ours) & 0.951 & 0.963 & \textbf{0.805} & 0.841 & 0.993 & 170 \\
Funnel-RankNet$^{\dagger}$ & \textbf{0.952} & \textbf{0.965} & \textbf{0.805} & \textbf{0.842} & \textbf{0.995} & 171 \\
Funnel-QBC$^{\dagger}$ & 0.733 & 0.639 & 0.432 & 0.632 & 0.550 & 242 \\
Retrieval-only$^{\dagger}$ & 0.871 & 0.826 & 0.700 & 0.774 & 0.841 & 228 \\
\addlinespace
\multicolumn{7}{l}{\textit{CUB-200-2011 (200 classes)}} \\
\system{} (ours) & 0.878 & 0.889 & \textbf{0.746} & \textbf{0.823} & \textbf{0.939} & 131 \\
Funnel-RankNet$^{\dagger}$ & \textbf{0.879} & \textbf{0.892} & 0.745 & \textbf{0.823} & \textbf{0.939} & 131 \\
Funnel-QBC$^{\dagger}$ & 0.849 & 0.844 & 0.470 & 0.785 & 0.619 & 143 \\
Retrieval-only$^{\dagger}$ & 0.855 & 0.867 & 0.742 & 0.816 & 0.875 & 135 \\
\addlinespace
\multicolumn{7}{l}{\textit{FGVC-Aircraft (90 classes)}} \\
\system{} (ours) & \textbf{0.727} & \textbf{0.737} & \textbf{0.564} & \textbf{0.588} & \textbf{0.838} & 256 \\
Funnel-RankNet$^{\dagger}$ & 0.718 & 0.730 & 0.563 & 0.585 & 0.834 & 257 \\
Funnel-QBC$^{\dagger}$ & 0.703 & 0.699 & 0.362 & 0.551 & 0.607 & 269 \\
Retrieval-only$^{\dagger}$ & 0.508 & 0.445 & 0.309 & 0.413 & 0.435 & 269 \\
\addlinespace
\multicolumn{7}{l}{\textit{iNaturalist 2021 (20 species)}} \\
\system{} (ours) & 0.581 & 0.698 & 0.606 & 0.539 & 0.818 & 602 \\
Funnel-RankNet$^{\dagger}$ & 0.551 & \textbf{0.706} & \textbf{0.613} & 0.513 & \textbf{0.822} & 602 \\
Funnel-QBC$^{\dagger}$ & \textbf{0.585} & 0.700 & 0.344 & \textbf{0.558} & 0.537 & 602 \\
Retrieval-only$^{\dagger}$ & 0.307 & 0.658 & 0.415 & 0.266 & 0.579 & 602 \\
\addlinespace
\multicolumn{7}{l}{\textit{fMoW (10 categories)}} \\
\system{} (ours) & \textbf{0.466} & 0.418 & 0.256 & 0.368 & 0.444 & 1022 \\
Funnel-RankNet$^{\dagger}$ & 0.459 & 0.425 & \textbf{0.261} & 0.368 & \textbf{0.454} & 1022 \\
Funnel-QBC$^{\dagger}$ & 0.453 & \textbf{0.429} & 0.160 & \textbf{0.375} & 0.283 & 1022 \\
Retrieval-only$^{\dagger}$ & 0.369 & 0.335 & 0.136 & 0.299 & 0.236 & 1022 \\
\bottomrule
\end{tabular}
\end{table}

\begin{table}[!t]
\centering
\caption{Mean rank of the ten methods of Tables~\ref{tab:results} and~\ref{tab:ablations} over the five datasets (fMoW as one dataset of 10 categories, as in Table~\ref{tab:results}), DINOv2.  On each dataset the methods are ranked by AULC-PosR (discovery) and by F1 (classifier), ties sharing the average rank (ranks use unrounded values).  Overall is the mean of the two.  Position 1 = best.  $^{\dagger}$~marks ablations of \system{}.}
\label{tab:meanrank}
\footnotesize
\begin{tabular}{l cccc cccc}
\toprule
 & \multicolumn{4}{c}{Clean labels} & \multicolumn{4}{c}{10\% label noise} \\
\cmidrule(lr){2-5}\cmidrule(lr){6-9}
Strategy & Pos. & Discovery & F1 & Overall & Pos. & Discovery & F1 & Overall \\
\midrule
\system{} (ours) & 1 & 1.80 & 3.00 & 2.40 & 2 & 3.00 & 2.00 & 2.50 \\
Funnel-RankNet$^{\dagger}$ & 2 & 2.00 & 3.40 & 2.70 & 2 & 2.00 & 3.00 & 2.50 \\
SEALS (most-likely-pos.) & 3 & 2.20 & 3.80 & 3.00 & 1 & 1.40 & 2.80 & 2.10 \\
PF-MA & 4 & 4.20 & 2.80 & 3.50 & 4 & 4.60 & 4.00 & 4.30 \\
Uncertainty & 5 & 6.60 & 3.80 & 5.20 & 7 & 7.60 & 6.00 & 6.80 \\
Funnel-QBC$^{\dagger}$ & 6 & 5.80 & 5.80 & 5.80 & 5 & 5.60 & 4.40 & 5.00 \\
Retrieval-only$^{\dagger}$ & 7 & 5.40 & 7.40 & 6.40 & 6 & 4.40 & 6.40 & 5.40 \\
SEALS & 8 & 8.80 & 7.20 & 8.00 & 9 & 9.00 & 9.00 & 9.00 \\
AnchorAL & 8 & 8.20 & 7.80 & 8.00 & 8 & 7.40 & 7.40 & 7.40 \\
Random & 10 & 10.00 & 10.00 & 10.00 & 10 & 10.00 & 10.00 & 10.00 \\
\bottomrule
\end{tabular}
\end{table}

\paragraph{Overall.}  \system{} has the best overall mean rank with clean labels (2.40).  It is first on discovery (1.80) and second on F1 (3.00, behind PF-MA at 2.80).  Funnel-RankNet (2.70) and SEALS (most-likely-pos.) (3.00) follow.  No method is best on every dataset.

\paragraph{The three benchmarks.}  Against SEALS, AnchorAL, Uncertainty and Random, \system{} is significantly ahead on every column of all three benchmarks.  On UC~Merced the F1 and Recall leads over SEALS, AnchorAL and Uncertainty come from the evaluation classifier: trained on labeled items alone, the gaps close (Appendix~\ref{app:evalclf}).  AULC-F1 then favors SEALS and Uncertainty, and \system{} keeps its lead only over AnchorAL (0.855 against 0.815).  The discovery leads (AULC-PosR, Pos.~Rec.) do not depend on the classifier.  Against PF-MA, no F1 difference is significant on the three benchmarks (0.011 or less), and \system{} discovers faster.  Its AULC-PosR is higher by 0.084, 0.172 and 0.112 on UC~Merced, CUB-200 and FGVC-Aircraft, on 21 of 21, 200 of 200 and 89 of 90 classes.  Against SEALS (most-likely-pos.) the result depends on the dataset.  On UC~Merced and CUB-200 no difference is significant, and SEALS (most-likely-pos.) is numerically ahead on AULC-PosR (0.824 against 0.805, and 0.756 against 0.746).  On FGVC-Aircraft \system{} is significantly ahead on all five columns, by 0.035 in F1 and 0.086 in AULC-PosR.  FGVC-Aircraft is the least separable benchmark (Figure~\ref{fig:separability}) and the one where retrieval alone does worst (Retrieval-only F1 0.508, Table~\ref{tab:ablations}).  Where every positive is found before the budget, the methods that seek positives stop early: on UC~Merced \system{}, SEALS (most-likely-pos.) and PF-MA use 170 to 173 of the 242 labels.

\paragraph{The large corpora.}  On iNaturalist \system{} discovers positives fastest of the methods in Table~\ref{tab:results} (AULC-PosR 0.606), significantly ahead of PF-MA, SEALS, AnchorAL and Uncertainty.  Uncertainty (0.726) and PF-MA (0.715) are significantly ahead of it on F1 (0.581).  No column differs significantly between SEALS (most-likely-pos.) and \system{}.  A wider retrieval neighborhood ($K{=}2{,}000$) raises \system{}'s F1 to 0.653 without changing its discovery (Appendix~\ref{app:k2000}).  On fMoW the leaders are close.  \system{} has the highest AULC-PosR in Table~\ref{tab:results} (0.256, against 0.254 for SEALS (most-likely-pos.)), ahead in 6 to 10 of the 10 categories against each external method.  Uncertainty and PF-MA have the highest F1 (0.468, against 0.466 for \system{}), and SEALS (most-likely-pos.) the highest Pos.~Rec.\ (0.451 against 0.444).

\paragraph{Ablations.}  No significant difference was detected on the benchmarks between \system{} and Funnel-RankNet, which never explores.  The largest difference is on FGVC-Aircraft, F1 0.727 against 0.718 (\system{} ahead on 56 of 90 classes, $p{=}3.1{\times}10^{-4}$).  On iNaturalist Funnel-RankNet discovers slightly faster (AULC-PosR 0.613 against 0.606, 18 of 20 classes, significant), and \system{} has the higher AULC-F1 (0.539 against 0.513, significant).  Removing the exploitation arm (Funnel-QBC) slows discovery on every dataset, by 0.20 to 0.37 in AULC-PosR on the benchmarks and 0.26 on iNaturalist, all significant, and in every fMoW class.  Removing the ranking stage (Retrieval-only) lowers F1 on all five datasets, most on FGVC-Aircraft (0.508) and iNaturalist (0.307).  On CUB-200 its discovery does not differ significantly from \system{}'s (AULC-PosR 0.742 against 0.746), and it finds fewer positives by the budget (Pos.~Rec.\ 0.875 against 0.939, significant).  The exploitation arm and the ranking stage carry \system{}'s results.  The exploration arm adds little that is measurable at the budget (Section~\ref{sec:discussion}).

\subsubsection{Computational Cost}\label{sec:cost}

We measured selection time in a dedicated run, separate from the tables: one task at a time on an otherwise idle machine (AMD EPYC 7V12 CPU, one NVIDIA A100 GPU, one thread per numerical library), 5 classes and 3 repeats per dataset, DINOv2, clean labels.  \system{} trains RankNet on the GPU, and every other learner runs on the CPU.  Every retrieving method uses exact search on the same index.  The times are for our implementations of the baselines.  Table~\ref{tab:cost} reports them.

\begin{table}[!t]
\centering
\caption{Selection time, dedicated timing run (5 classes, 3 repeats per dataset, DINOv2, clean labels, one task at a time).  Seconds per round and per task (the whole selection of one class and repeat).  GAL was not run on iNaturalist.}
\label{tab:cost}
\footnotesize
\begin{tabular}{l cc cc}
\toprule
 & \multicolumn{2}{c}{FGVC-Aircraft (8{,}000 images)} & \multicolumn{2}{c}{iNaturalist (2.15M images)} \\
\cmidrule(lr){2-3}\cmidrule(lr){4-5}
Strategy & s / round & s / task & s / round & s / task \\
\midrule
\system{} (ours)         & 0.137 & 1.59 & 3.97 & 105 \\
SEALS (most-likely-pos.) & 0.146 & 1.75 & 29.2 & 876 \\
PF-MA                    & 0.011 & 0.13 & 2.25 & 67.5 \\
SEALS                    & 0.155 & 1.86 & 28.7 & 862 \\
AnchorAL                 & 0.103 & 1.24 & 27.9 & 837 \\
Uncertainty              & 0.025 & 0.30 & 5.02 & 151 \\
Random                   & 0.000 & 0.000 & 0.000 & 0.003 \\
GAL (batch 20)           & 34.1 & 409 & -- & -- \\
\bottomrule
\end{tabular}
\end{table}

On FGVC-Aircraft every method except GAL selects a whole task in under 2\,s, which has no practical weight next to the annotator's time.  GAL takes 409\,s per task, about 260 times \system{}, because it refits a linear SVM for every candidate and pseudo-label at each batch slot.  On iNaturalist the differences are large.  \system{} selects a task in 105\,s, eight times less than the three neighborhood baselines (837 to 876\,s) and 1.4 times less than Uncertainty, which scores the whole pool every round.  PF-MA, which fits one linear SVM and scores the pool, is cheaper still (67.5\,s).  Almost all of \system{}'s time is retrieval (101 of 105\,s), and the ranking stage adds 3.4\,s per task.  \system{} issues one batched search per round with every labeled positive as a query, while the neighborhood baselines, as we implemented them, look up the neighbors of each selected item or anchor separately, with at least two search calls per lookup.  Part of this gap is therefore implementation.  AnchorAL's per-anchor searches could be issued as one batch, while SEALS adds the neighbors of each pick before the next pick by design.  \system{} also needs fewer rounds (11.5 and 26.6 against 12 and 30), because its batch grows after all-positive batches.  We did not time RankNet on a CPU.

\subsubsection{Trigger Timing and Batch Growth}\label{sec:ablation}

\system{} has two adaptive components: the precision trigger that starts the mixed phase, and the batch growth after all-positive batches.  Table~\ref{tab:ablation} replaces each with a fixed rule on the three benchmarks, everything else unchanged.  The mixture can start in the first round in which RankNet can train (two positives and two negatives labeled), or at a fixed round (3, 5 or 8, the warm-up round counted).  Fixed~$k$ keeps the batch at 20.  The count trigger starts the mixed phase once 30 positives are labeled and also keeps the batch at 20, so it differs from Fixed~$k$ only in the trigger.

\begin{table}[t]
\centering
\caption{Replacing \system{}'s two adaptive components with fixed rules, DINOv2, clean labels, read as Table~\ref{tab:results}.  The switch rules change only when the mixed phase starts.  Fixed~$k$ changes only the batch size.  The count trigger row was run on a second machine, together with \system{} (within 0.001 of the first row of each block) and Fixed~$k$, which the text compares it with.  Rounds = mean annotation rounds per task.}
\label{tab:ablation}
\footnotesize
\begin{tabular}{l cccccc}
\toprule
Variant & F1 & Recall & AULC-PosR & AULC-F1 & Pos.\ Rec. & Rounds \\
\midrule
\addlinespace
\multicolumn{7}{l}{\textit{UC Merced}} \\
\system{} (precision trigger) & 0.951 & 0.963 & 0.805 & 0.841 & 0.993 & 7.5 \\
Mixture from the first trainable round & 0.952 & 0.965 & 0.785 & 0.827 & 0.994 & 7.9 \\
Switch at round 3 & 0.952 & 0.965 & 0.790 & 0.831 & 0.994 & 7.8 \\
Switch at round 5 & 0.952 & 0.963 & 0.805 & 0.841 & 0.993 & 7.5 \\
Switch at round 8 & 0.952 & 0.966 & 0.805 & 0.842 & 0.994 & 7.6 \\
Fixed $k$ (no batch growth) & 0.951 & 0.963 & 0.813 & 0.845 & 0.993 & 8.0 \\
Count trigger (30 positives), fixed $k$ & 0.952 & 0.965 & 0.806 & 0.842 & 0.993 & 8.1 \\
\addlinespace
\multicolumn{7}{l}{\textit{CUB-200-2011}} \\
\system{} (precision trigger) & 0.878 & 0.889 & 0.746 & 0.823 & 0.939 & 6.6 \\
Mixture from the first trainable round & 0.877 & 0.886 & 0.728 & 0.821 & 0.939 & 6.7 \\
Switch at round 3 & 0.878 & 0.888 & 0.736 & 0.822 & 0.939 & 6.7 \\
Switch at round 5 & 0.877 & 0.888 & 0.745 & 0.823 & 0.938 & 6.6 \\
Switch at round 8 & 0.879 & 0.891 & 0.745 & 0.823 & 0.939 & 6.7 \\
Fixed $k$ (no batch growth) & 0.879 & 0.889 & 0.752 & 0.824 & 0.940 & 6.9 \\
Count trigger (30 positives), fixed $k$ & 0.878 & 0.888 & 0.752 & 0.824 & 0.939 & 6.9 \\
\addlinespace
\multicolumn{7}{l}{\textit{FGVC-Aircraft}} \\
\system{} (precision trigger) & 0.727 & 0.737 & 0.564 & 0.588 & 0.838 & 12.2 \\
Mixture from the first trainable round & 0.723 & 0.732 & 0.558 & 0.589 & 0.838 & 12.5 \\
Switch at round 3 & 0.723 & 0.732 & 0.557 & 0.588 & 0.837 & 12.5 \\
Switch at round 5 & 0.728 & 0.737 & 0.564 & 0.588 & 0.837 & 12.3 \\
Switch at round 8 & 0.725 & 0.734 & 0.564 & 0.586 & 0.840 & 12.1 \\
Fixed $k$ (no batch growth) & 0.727 & 0.737 & 0.565 & 0.588 & 0.838 & 12.6 \\
Count trigger (30 positives), fixed $k$ & 0.723 & 0.733 & 0.561 & 0.586 & 0.835 & 12.6 \\
\bottomrule
\end{tabular}
\end{table}

\emph{Starting the mixture early slows discovery, and F1 at the budget changes by at most 0.005.}  From the first trainable round, AULC-PosR is lower than with the precision trigger by 0.020 on UC~Merced (20 of 21 classes, $p{=}6.7{\times}10^{-6}$) and by 0.017 on CUB-200 (178 of 200 classes, $p{=}3.2{\times}10^{-25}$), and by 0.006 on FGVC-Aircraft ($p{=}0.007$, not significant).  A switch at round 3 lowers it by 0.015, 0.009 and 0.007, all significant.  F1 stays within 0.005 of the trigger's in every case.  Between a switch at round 5 or 8 and the precision trigger, no significant difference was detected on any column.  Runs that end before the fixed round never mix: at round 8 these are 64\%, 54\% and 8\% of the tasks, at round 5 17\% of the CUB-200 tasks.  The trigger fires at round 5 in 69\% to 82\% of the tasks.  Its window needs three exploitation batches after the warm-up, so round 5 is the earliest round at which it can fire.  In most tasks the trigger therefore fires at the earliest round it can, the same round as a fixed switch at round 5.  So this test cannot tell whether the trigger helps by reading precision or simply by waiting about four rounds.  What it does show is that exploration should wait until the exploitation arm has run for a few rounds.  The trigger finds this point from the observed precision, with no round number to tune.  On these benchmarks no significant difference was detected between the trigger and a fixed switch at round 5 or 8.  The precision trigger is one workable policy, and its added benefit in the tested settings is limited.

\emph{No significant difference was detected between a count rule and the precision trigger.}  The count trigger uses 30 positives, the code's default threshold, with the same warm-up as \system{}.  The mixed phase starts at round 4 on UC~Merced and CUB-200 and at round 5 on FGVC-Aircraft (medians).  Against a Fixed~$k$ run on the same machine, which differs from it only in the trigger, F1 is within 0.005 on every benchmark and no column differs significantly.  AULC-PosR is lower on UC~Merced (0.806 against 0.813) and FGVC-Aircraft (0.561 against 0.566), with $p{=}6.1{\times}10^{-4}$ and $0.003$ (not significant), and the same on CUB-200 (0.752).  So on these benchmarks no significant difference was detected between the precision trigger and the count trigger with the batch fixed at 20, and the precision trigger needs no threshold on the number of positives.

\emph{Batch growth saves annotation rounds.}  With a fixed batch of 20, F1 at the budget differs by at most 0.001, AULC-PosR is slightly higher (0.813 against 0.805 on UC~Merced, 0.752 against 0.746 on CUB-200), and runs take more rounds (8.0, 6.9 and 12.6 against 7.5, 6.6 and 12.2).  Growth thus spares the annotator retrain-and-wait cycles, at a small cost in discovery speed (AULC-PosR 0.006 to 0.008 lower on UC~Merced and CUB-200).

\subsubsection{Robustness to Annotator Label Noise}\label{sec:noise}

\begin{figure*}[t]
  \centering
  \includegraphics[width=\textwidth]{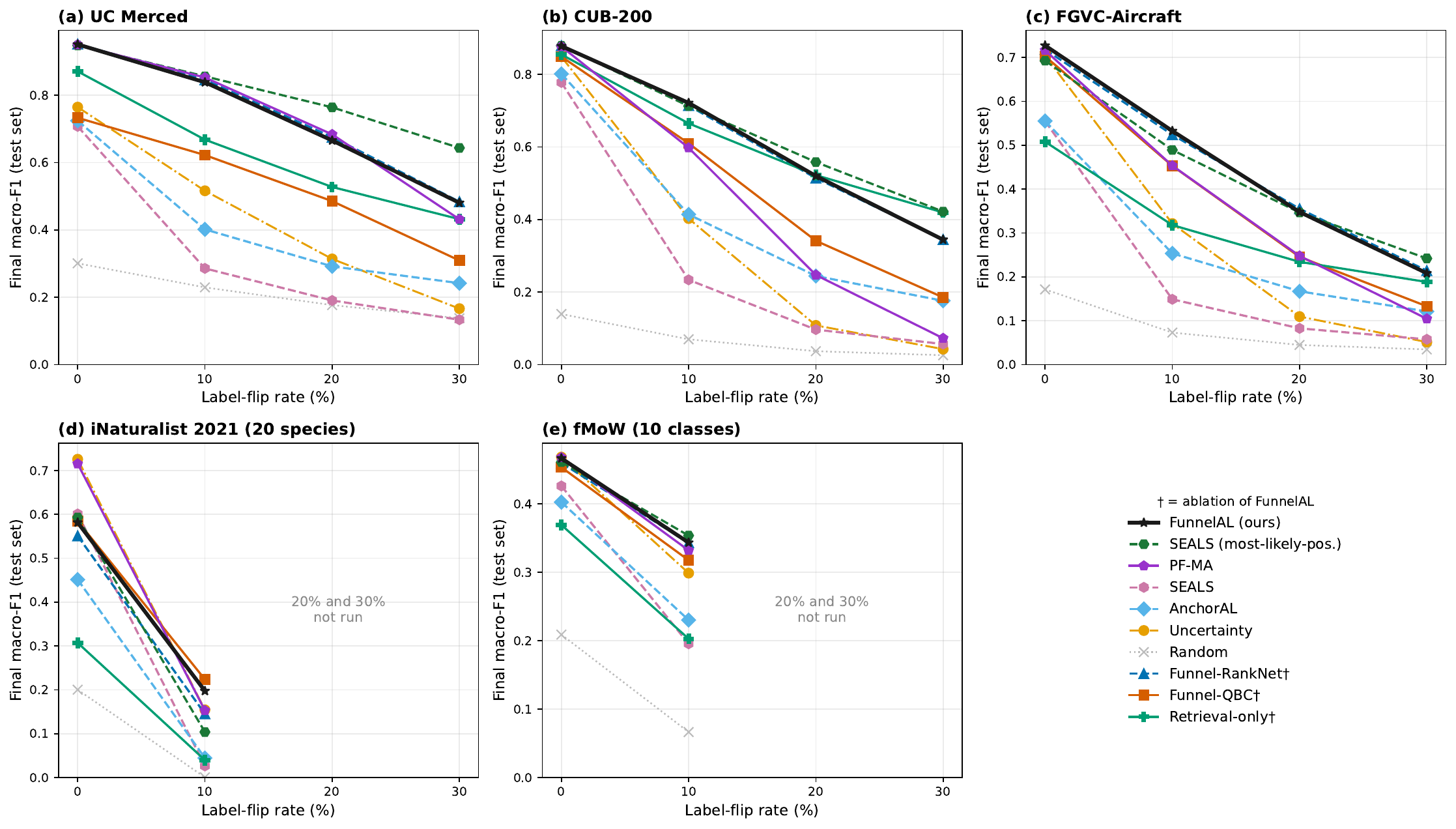}
  \caption{F1 on the test set against the label-flip rate, DINOv2, the ten methods of Table~\ref{tab:meanrank}, values at the budget, mean over 5 repeats and then over classes.  All five datasets are run at 0\% and 10\%, and the three benchmarks also at 20\% and 30\%.  The 20\% and 30\% points come from an earlier run of our code.  At 10\% that run's records match the reported runs task by task.  fMoW's panel pools its ten categories, as the tables do.  $^{\dagger}$~marks ablations of \system{}.}
  \Description{Five line charts, one per dataset, showing final F1 against the label-flip rate for ten strategies.}
  \label{fig:noise}
\end{figure*}

Real-world databases are estimated to hold about 5\% encoding errors when no specific measures are taken~\citep{frenay2014label}, and reported label-noise rates in real-world image datasets range from 8\% to 38.5\%~\citep{song2022noisy}.  We rerun the protocol with a symmetric-noise annotator~\citep{frenay2014label}, in which both error directions are equally likely: every queried label is flipped independently with probability $p$.  All five datasets are run at $p{=}10\%$, and the three benchmarks also at 20\% and 30\%.  The corrupted labels drive \emph{everything the system observes}: the labeled sets, the evaluation classifier's training data, and the batch precision that controls \system{}'s growth rule and trigger.  Apart from the clean seed pair, the stop rule and the benchmark budgets (Section~\ref{sec:experiments}), ground truth is used only for measurement.  Under noise, Pos.~Rec.\ counts only true positives labeled positive.  Table~\ref{tab:noise} reports the values at 10\%, Figure~\ref{fig:noise} traces F1 across the rates, and Table~\ref{tab:meanrank} gives the mean rank at 10\%.

\begin{table}[!t]
\centering
\caption{External methods at 10\% label noise, DINOv2, read as Table~\ref{tab:results}.  Pos.~Rec.\ counts true positives labeled positive.}
\label{tab:noise}
\footnotesize
\begin{tabular}{l cccccc}
\toprule
Strategy & F1 & Recall & AULC-PosR & AULC-F1 & Pos.\ Rec. & Labels \\
\midrule
\addlinespace
\multicolumn{7}{l}{\textit{UC Merced (21 classes)}} \\
\system{} (ours) & 0.839 & 0.883 & 0.696 & 0.752 & 0.878 & 202 \\
SEALS (most-likely-pos.) & \textbf{0.856} & \textbf{0.901} & \textbf{0.727} & \textbf{0.774} & \textbf{0.890} & 172 \\
PF-MA & 0.853 & 0.896 & 0.611 & 0.672 & 0.887 & 187 \\
SEALS & 0.286 & 0.254 & 0.127 & 0.211 & 0.188 & 242 \\
AnchorAL & 0.402 & 0.386 & 0.233 & 0.355 & 0.307 & 242 \\
Uncertainty & 0.516 & 0.509 & 0.247 & 0.375 & 0.415 & 242 \\
Random & 0.229 & 0.209 & 0.081 & 0.150 & 0.141 & 242 \\
\addlinespace
\multicolumn{7}{l}{\textit{CUB-200-2011 (200 classes)}} \\
\system{} (ours) & \textbf{0.721} & 0.798 & 0.621 & \textbf{0.710} & 0.790 & 138 \\
SEALS (most-likely-pos.) & 0.712 & \textbf{0.822} & \textbf{0.652} & 0.700 & \textbf{0.825} & 130 \\
PF-MA & 0.598 & 0.724 & 0.331 & 0.433 & 0.691 & 141 \\
SEALS & 0.234 & 0.308 & 0.059 & 0.173 & 0.094 & 143 \\
AnchorAL & 0.414 & 0.505 & 0.170 & 0.402 & 0.228 & 143 \\
Uncertainty & 0.403 & 0.493 & 0.101 & 0.275 & 0.194 & 143 \\
Random & 0.070 & 0.107 & 0.029 & 0.074 & 0.035 & 143 \\
\addlinespace
\multicolumn{7}{l}{\textit{FGVC-Aircraft (90 classes)}} \\
\system{} (ours) & \textbf{0.533} & \textbf{0.601} & \textbf{0.426} & \textbf{0.447} & \textbf{0.638} & 264 \\
SEALS (most-likely-pos.) & 0.489 & 0.575 & 0.386 & 0.408 & 0.591 & 264 \\
PF-MA & 0.453 & 0.551 & 0.282 & 0.304 & 0.563 & 266 \\
SEALS & 0.149 & 0.216 & 0.049 & 0.118 & 0.080 & 269 \\
AnchorAL & 0.253 & 0.313 & 0.114 & 0.218 & 0.175 & 269 \\
Uncertainty & 0.321 & 0.393 & 0.100 & 0.213 & 0.211 & 269 \\
Random & 0.073 & 0.140 & 0.028 & 0.064 & 0.041 & 269 \\
\addlinespace
\multicolumn{7}{l}{\textit{iNaturalist 2021 (20 species)}} \\
\system{} (ours) & \textbf{0.197} & 0.604 & 0.464 & \textbf{0.311} & 0.653 & 602 \\
SEALS (most-likely-pos.) & 0.104 & \textbf{0.619} & \textbf{0.485} & 0.167 & \textbf{0.658} & 602 \\
PF-MA & 0.153 & 0.476 & 0.250 & 0.114 & 0.497 & 602 \\
SEALS & 0.026 & 0.381 & 0.020 & 0.044 & 0.034 & 602 \\
AnchorAL & 0.045 & 0.463 & 0.078 & 0.079 & 0.118 & 602 \\
Uncertainty & 0.154 & 0.257 & 0.032 & 0.098 & 0.073 & 602 \\
Random & 0.001 & 0.192 & 0.005 & 0.007 & 0.005 & 602 \\
\addlinespace
\multicolumn{7}{l}{\textit{fMoW (10 categories)}} \\
\system{} (ours) & 0.343 & 0.322 & 0.174 & 0.263 & 0.309 & 1022 \\
SEALS (most-likely-pos.) & \textbf{0.353} & \textbf{0.356} & \textbf{0.190} & \textbf{0.276} & \textbf{0.342} & 1022 \\
PF-MA & 0.332 & 0.339 & 0.118 & 0.253 & 0.257 & 1022 \\
SEALS & 0.195 & 0.295 & 0.029 & 0.146 & 0.054 & 1022 \\
AnchorAL & 0.230 & 0.302 & 0.049 & 0.193 & 0.087 & 1022 \\
Uncertainty & 0.299 & 0.287 & 0.062 & 0.223 & 0.131 & 1022 \\
Random & 0.067 & 0.274 & 0.011 & 0.060 & 0.019 & 1022 \\
\bottomrule
\end{tabular}
\end{table}

\emph{At 10\% noise.}  \system{} has the highest F1 of the methods in Table~\ref{tab:noise} on CUB-200, FGVC-Aircraft and iNaturalist.  \system{}'s F1 lead is significant on FGVC-Aircraft, over SEALS (most-likely-pos.) by 0.043 and over PF-MA by 0.079, and on iNaturalist over SEALS (most-likely-pos.), by 0.094.  On CUB-200 its lead over SEALS (most-likely-pos.) is not significant.  On UC~Merced, SEALS (most-likely-pos.) (0.856) and PF-MA (0.853) are numerically ahead of \system{} (0.839, not significant), and on fMoW SEALS (most-likely-pos.) is ahead (0.353 against 0.343, in 7 of the 10 categories).  Among the ablations, Funnel-QBC has a higher F1 than \system{} on iNaturalist (0.224 against 0.197, not significant, Appendix~\ref{app:noise-ablations}).  On discovery, SEALS (most-likely-pos.) is the stronger method on two benchmarks.  Its AULC-PosR is significantly higher on UC~Merced (by 0.031, 19 of 21 classes) and CUB-200 (by 0.032, 158 of 200 classes), and \system{}'s is significantly higher on FGVC-Aircraft (by 0.040).  SEALS (most-likely-pos.) therefore has the best mean rank at 10\% noise (2.10), with \system{} and Funnel-RankNet tied second (2.50).

\emph{At higher rates.}  On the three benchmarks the methods that sample uncertain items lose the most.  At 30\% noise Uncertainty keeps 22\%, 5\% and 7\% of its noise-free F1 (UC~Merced, CUB-200, FGVC-Aircraft), SEALS 19\%, 7\% and 10\%, and PF-MA 45\%, 8\% and 15\%.  \system{} keeps 51\%, 39\% and 29\%.  SEALS (most-likely-pos.) keeps 68\%, 48\% and 35\%, and at 30\% it has the highest F1 of all methods on all three benchmarks.  At 20\% and 30\% we report means only and ran no significance tests.  On iNaturalist every method loses most of its F1 already at 10\%.  \system{} keeps 34\%, more than any method in Table~\ref{tab:noise} (Random 0.3\%, the others 4\% to 21\%).  A plausible reading of the pattern, not tested here, is that a flipped label does the most damage where a method samples near its decision boundary, and the least where it samples confident positives whose neighbors carry clean labels.

Our noise model is uniform and symmetric.  Real annotator errors concentrate on ambiguous samples~\citep{frenay2014label}, which this model does not capture (Section~\ref{sec:limitations}).

\subsubsection{A Second Encoder}\label{sec:clip}

We repeated the comparison of the ten methods on all five datasets with CLIP ViT-B/16, with clean labels and at 10\% noise (Appendix~\ref{app:clip}).  CLIP's embedding separates the fine-grained classes less well: every method's F1 drops on CUB-200 (best 0.754 against 0.879 with DINOv2) and FGVC-Aircraft (0.467 against 0.727).  With clean labels the order of the methods changes little.  \system{} and Funnel-RankNet share the best mean rank (2.40), ahead of PF-MA (3.20) and SEALS (most-likely-pos.) (3.50).  At 10\% noise the top changes: Funnel-RankNet is first (1.50) and \system{} second (1.70), ahead of SEALS (most-likely-pos.) (3.30), which is first with DINOv2.  With clean labels \system{} discovers significantly faster than SEALS (most-likely-pos.) on CUB-200, FGVC-Aircraft and iNaturalist, and, as with DINOv2, Uncertainty has the higher F1 on iNaturalist (0.517 against 0.361).

\section{Discussion}\label{sec:discussion}

\subsection{Cross-Dataset Analysis}

\paragraph{(1) What the funnel bounds is the ranker's candidate set, not final quality.}
The retrieval stage limits how many candidates the ranking stage scores, whatever the corpus size.  Retrieval itself is not bounded.  Exact search compares every labeled positive with every image, so its cost grows with the corpus and with the number of positives, and on iNaturalist it takes 101 of \system{}'s 105\,s per task.  With our implementations, \system{}'s selection there is eight times faster than the neighborhood baselines and 1.4 times faster than full-pool uncertainty sampling (Table~\ref{tab:cost}).  PF-MA, which fits one linear SVM per round, is faster still.  We do not claim that the funnel improves final quality.  On iNaturalist, PF-MA and uncertainty sampling reach a higher F1 under our evaluation protocol (Uncertainty 0.726, PF-MA 0.715, \system{} 0.581), and retrieval alone is weak where classes are entangled (Retrieval-only F1 0.508 on FGVC-Aircraft and 0.307 on iNaturalist, Table~\ref{tab:ablations}).  Retrieval narrows the search, and the ranking stage carries the quality.

\paragraph{(2) Discovery and classifier quality are different goals.}
The methods that seek likely positives (\system{}, SEALS (most-likely-pos.), Funnel-RankNet) lead on discovery, with mean discovery ranks of 1.80 to 2.20 (Table~\ref{tab:meanrank}).  PF-MA, which ranks the most ambiguous predicted positives first, reaches an F1 not significantly different from \system{}'s on the benchmarks, and PF-MA and Uncertainty reach a higher F1 on iNaturalist.  On the benchmarks the two goals coincide for \system{}: no significant F1 difference from PF-MA was detected, and it discovers faster (AULC-PosR higher by 0.08 to 0.17).  On iNaturalist's working set of 2.15 million images they part.  Within 600 labels, the methods that label the neighborhoods of found positives find the class sooner, and the methods that label near the boundary reach a higher F1.  A wider retrieval neighborhood narrows this gap: with $K{=}2{,}000$ \system{}'s F1 on iNaturalist rises to 0.653 while its discovery stays the same (Appendix~\ref{app:k2000}).  For an annotation tool, whose output is the labeled set, we consider discovery the primary goal, and we report both.

\paragraph{(3) Where \system{} leads and where it does not.}
With DINOv2 and clean labels, \system{} is significantly ahead of SEALS, AnchorAL, Uncertainty and Random on every column of the three benchmarks.  It also discovers faster than PF-MA (AULC-PosR) on every dataset, significantly on the four tested ones and in 7 of the 10 fMoW categories.  On UC~Merced its leads in classifier quality depend on the evaluation classifier (Appendix~\ref{app:evalclf}).  Its closest competitor is SEALS's active-search mode, which also seeks likely positives inside a similarity neighborhood.  With clean labels no significant difference between the two was detected on UC~Merced, CUB-200 and iNaturalist.  \system{} is ahead on FGVC-Aircraft, the least separable benchmark, on all five columns.  With DINOv2 and noisy labels SEALS (most-likely-pos.) is the more robust method: it discovers faster at 10\% noise on UC~Merced and CUB-200, and it has the highest F1 at 30\% on all three benchmarks (Section~\ref{sec:noise}).  With CLIP at 10\% noise \system{} discovers faster than it on CUB-200 (Appendix~\ref{app:clip}).

\begin{figure*}[t]
  \centering
  \includegraphics[width=\textwidth]{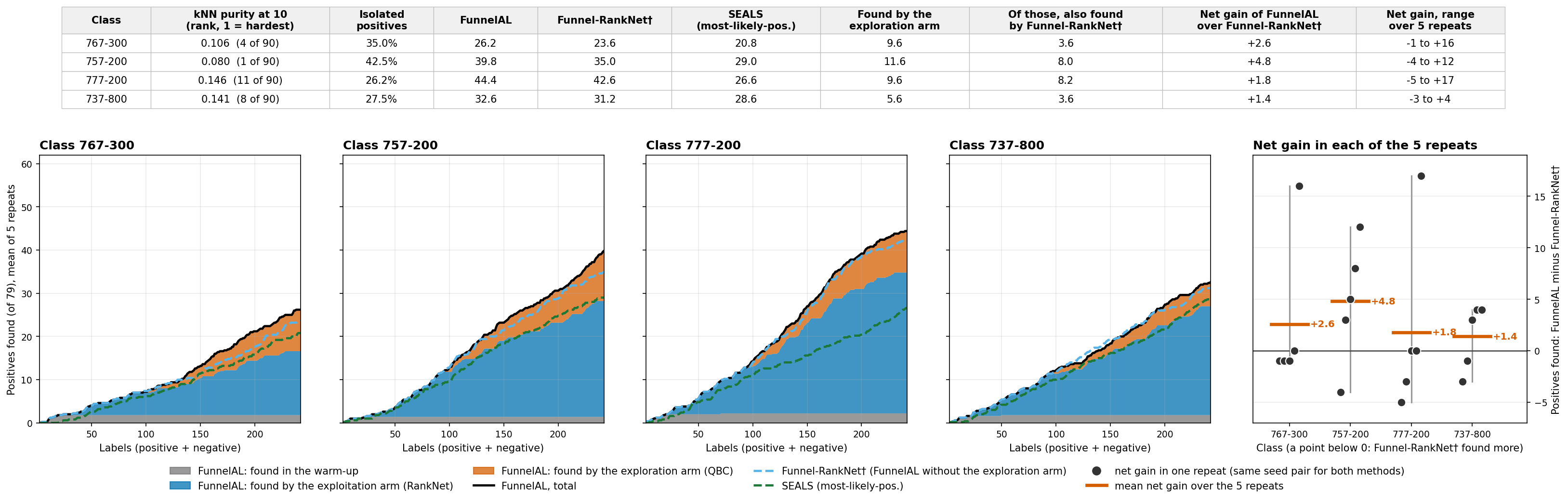}
  \caption{Positives found against labels on four FGVC-Aircraft classes (the Boeing variants of Figure~\ref{fig:tsne}), DINOv2, clean labels, 79 positives to find and a budget of 240 labels per class, mean of 5 repeats.  The runs are a rerun of those of Table~\ref{tab:results} that records which arm proposed each label, on a second machine, where 379 of 450 \system{} tasks match the reported runs exactly.  \system{}'s positives are split by the arm that proposed them (warm-up, exploitation, exploration).  Funnel-RankNet$^{\dagger}$ (without the exploration arm) and SEALS (most-likely-pos.) are shown as lines.  The table gives each class's kNN purity at 10, the share of a positive's 10 nearest neighbors in the same class, with its rank among the 90 classes on the working set.  Isolated positives are those with no image of their class among their 10 nearest neighbors, and the table gives their share.  It also gives the positives found within the budget, those found by the exploration arm, and how many of those Funnel-RankNet also finds.  The right panel shows \system{}'s net gain over Funnel-RankNet in each repeat.  Five repeats per class, no test.}
  \Description{Four area plots of positives found against labels for four aircraft classes, split by arm, with a table above and a panel of per-repeat net gains.}
  \label{fig:arm-capture}
\end{figure*}

\paragraph{(4) Where the exploration arm's positives come from.}
On the three benchmarks with DINOv2 and clean labels, starting exploration at the first trainable round or at round 3 slows discovery.  No significant difference was detected between the precision trigger and a fixed switch at round 5 or 8, or, with the batch fixed at 20, a switch after 30 positives (Section~\ref{sec:ablation}).  The precision trigger is one workable policy, and its added benefit in the tested settings is limited.  At the budget, the exploration arm adds little that is measurable: no significant difference was detected on the benchmarks between \system{} and Funnel-RankNet, which never explores (Table~\ref{tab:ablations}).  Figure~\ref{fig:arm-capture} follows the four Boeing variants highlighted in Figure~\ref{fig:tsne}, chosen there to show how entangled the FGVC-Aircraft variants are.  No method finds more than 57\% of their positives within the budget, and on 767-300 and 737-800 none finds half.  The exploration arm supplies 17 to 37\% of \system{}'s positives in these classes.  Most of them are not unique to it: Funnel-RankNet, without exploration, finds 38 to 85\% of the same images within the budget.  In this rerun \system{} finds 1.4 to 4.8 more positives per run than Funnel-RankNet, with swings from $-5$ to $+17$ between repeats (1.4 to 3.8 in the runs of Table~\ref{tab:results}).  Over all 90 classes the mixed phase gives no gain in positive recall.  The figure shows how the work is divided between the arms, not that exploration adds positives.

\subsection{Limitations and Future Work}\label{sec:limitations}

\emph{The difficulty of the task itself.}
Finding one class in a large unlabeled corpus from one positive and one negative example is hard for every method.  On iNaturalist a species makes up about 0.01\% of 2.15 million images, so the first retrieval step sees 511 candidates around a single seed, and in 3 of 100 tasks \system{} never finds a second positive within the budget, so once that neighborhood is used up it labels random images.  Every method gets the same 1+1 start, while the baselines' papers start from 6 to 100 labeled items.  At 10\% label noise every method loses most of its F1 on iNaturalist.  In real use an annotator controls several things that our simulation fixes.  They choose the seed images, widen or narrow the retrieval step, add a new seed when a search stops finding positives, and decide when to stop.  Our benchmark gives no method this help.  The seeds are random and the other settings are fixed.

\emph{Simulated annotator and user study.}
Section~\ref{sec:noise} tests uniform symmetric label noise, but real annotator errors are \emph{instance-dependent}~\citep{frenay2014label}: they concentrate on ambiguous, boundary-proximate samples (especially those selected by the QBC component), so mislabeling rates on funnel-recommended batches may exceed the uniform rates we simulate, and difficulty-weighted noise models remain to be tested.  Real annotators also exhibit fatigue over long sessions and vary in expertise.  Furthermore, the simulated annotator does not model \emph{time}: some recommended samples are easier or harder to judge, and a time-aware cost model could change the optimal explore-exploit balance.  The most important next step is a controlled user study with domain experts (e.g., remote sensing analysts) to validate that (a)~funnel recommendations are perceived as useful, (b)~user-controlled exploration-exploitation improves on the automated precision-triggered policy, and (c)~wall-clock annotation time savings match the label-count savings.

\emph{Ground truth in the protocol.}
A run ends when every positive in the working set has been queried, and the benchmark budgets are three times the number of positives.  Both read the ground truth, which a deployed system does not have.  Values are read at the budget, so a run that stopped early keeps its last value.  A deployed system needs a principled stopping criterion.  One direction is to estimate the fraction of positives already discovered from the rate at which new positives appear over recent rounds.  A sustained decline could signal near-exhaustive annotation, while a sudden increase (e.g., discovering a new cluster) could indicate significant remaining positives.

\emph{Evaluation classifier.}
The evaluation classifier treats a random sample of unlabeled items as negatives.  Positives a method has not found enter that sample, which can affect the F1 of methods that leave many positives unfound.  We measured the effect on UC~Merced only (Appendix~\ref{app:evalclf}), and it remains unmeasured on the other datasets.

\emph{Seed selection and distribution mismatch.}
When the positive seed comes from an external source, its feature distribution may differ from that of the target class within the corpus (e.g., a ground-level reference photo vs.\ overhead satellite imagery of the same object).  This mismatch can weaken the initial retrieval stage, since embedding similarity between the external seed and in-corpus positives may be low.  Our experiments use in-corpus seeds.  Investigating robustness to out-of-distribution seeds is an important practical direction.

\emph{Fixed settings.}
We use $K{=}512$, $C{=}5{,}000$, $\tau{=}0.7$, $w{=}3$, a 60/40 split and $\gamma{=}1.2$ on every dataset.  Changing $\tau$, $w$, the split or $\gamma$ one at a time moves F1 and AULC-PosR by at most 0.012 with clean labels on CUB-200, FGVC-Aircraft and iNaturalist (Appendix~\ref{app:sensitivity}).  At 10\% noise on iNaturalist the split trades one goal against the other: a larger RankNet share raises AULC-PosR and lowers F1 (by 0.037 at 80\%).  Retrieval breadth matters more at scale: with $K{=}2{,}000$ \system{}'s F1 on iNaturalist rises by 0.07 (Appendix~\ref{app:k2000}).  Retrieval parameters should therefore scale with the corpus size and the expected class prevalence, and the ranking parameters could be learned online using contextual bandits~\citep{li2010contextual, chapelle2011empirical}, treating the choice of ranking strategy as an arm selection problem with the reward defined by the classifier's improvement.

\emph{Scale.}
The largest set we search is iNaturalist's working set of 2.15 million images.  We search it exactly, comparing each query with every image.  Corpora a hundred times larger or more would likely need approximate nearest-neighbor search or a coarse-to-fine candidate search.  Both can miss true neighbors, and we did not test how that changes discovery.

\emph{Encoders.}
Both encoders are general-purpose vision transformers.  With clean labels the methods fall in a similar order under both (Appendix~\ref{app:clip}), but domain-specific encoders, for example for medical or remote-sensing images, remain untested.

\emph{Other data modalities.}
\system{} never inspects the images themselves.  Every stage operates on the embedding vectors: retrieval on $\ell_2$-normalized vectors, RankNet on the raw vectors, and the committee and the evaluation classifier on standardized vectors.  The only image-specific component is the frozen encoder that produces them.  The system should therefore transfer to any data type with a strong pretrained encoder, e.g., text passages, audio clips, or protein sequences, by swapping the encoder.  Our evaluation covers only images, so this generality remains to be demonstrated, and modalities whose embedding geometry differs from vision encoders' (for instance, the anisotropic spaces of some text-embedding models, where similarity scores concentrate in a narrow range) may need retuned retrieval parameters.

\section{Conclusion}\label{sec:conclusion}

We introduced \system{}, a retrieve-then-rank active learning system for single-class discovery that recasts data annotation as a multi-stage funnel recommendation problem.  Embedding-based retrieval (DWVA) narrows the corpus to a bounded candidate set, and a ranking stage exploits a learned ranker (RankNet) and later mixes in committee-based exploration (QBC).  We evaluated it on three benchmarks and two large corpora of up to 2.7 million images, with two encoders.  With clean labels and DINOv2, \system{} has the best mean rank over the five datasets.  It discovers positives faster than every external method except SEALS's active-search mode.  No significant difference from that mode was detected on most datasets, and \system{} leads it on the least separable benchmark.

Three conclusions follow.  First, the funnel bounds the number of candidates the ranker scores: on iNaturalist's working set of 2.15 million images \system{} selects eight times faster than the neighborhood baselines as we implemented them.  Second, discovery and classifier quality are different goals at scale: on iNaturalist, PF-MA and uncertainty sampling reach the higher F1 while \system{} finds the class sooner.  Third, exploration should not start too early.  On the three benchmarks with DINOv2 and clean labels, starting it at the first trainable round or at round 3 slows discovery.  No significant difference was detected between the precision trigger and a switch at round 5 or 8, or, with the batch fixed at 20, a switch after 30 positives.  The trigger is one workable policy whose added benefit in these settings is limited.  Under 10\% label noise \system{} has a higher F1 than every baseline on three of the five datasets with DINOv2 and on all five with CLIP, and SEALS's active-search mode is the more robust at higher noise rates.  We believe these results make a case for closer integration between the recommender systems and active learning communities, and suggest that \system{} can serve as a concrete starting point for that dialogue.

\subsubsection*{Broader Impact Statement}

\system{} aims to lower the human cost of finding and labeling the instances of a single class of interest in a large corpus.  This matters most in domains where the labeling budget, not the algorithm, is the bottleneck.  In ecological surveys, medical imaging, and remote sensing, expert time is the scarce resource, and \system{} could make those hours go further.  Our results come from simulated annotators, so these savings are prospective.  Reported label-noise rates in real-world image datasets range from 8 to 38.5\%.  At 10\% label noise \system{} has a higher F1 than every baseline on three of our five datasets with DINOv2 and on all five with CLIP.  At higher rates SEALS's active-search mode is more robust, and every method loses much of its accuracy on the largest corpus.

Users should also know that \system{} can miss parts of a class.  It finds samples through embedding similarity, so sub-populations that the embedding represents poorly are found late or not at all.  In our experiments no method finds more than 57\% of the positives of four entangled FGVC-Aircraft variants within the budget (Figure~\ref{fig:arm-capture}).  A dataset built this way can look complete while missing sub-populations, and estimating how much was missed is itself an open problem.

\bibliographystyle{tmlr}
\bibliography{references}

\appendix

\section{Retrieval Breadth on iNaturalist}\label{app:k2000}

With $K{=}512$ the first candidate set on iNaturalist holds 511 of 2.15 million images.  We reran \system{} with $K{=}2{,}000$ and $C{=}20{,}000$, everything else unchanged.  F1 at the budget rose from 0.581 to 0.653 (higher on 16 of 20 species, $p{=}0.002$, not significant at our threshold).  The gain comes from precision.  On the test set the evaluation classifier's precision rose from 0.548 to 0.646, while its recall stayed near 0.70 (0.698 to 0.692).  Discovery did not change (Pos.~Rec.\ 0.818 to 0.816, AULC-PosR 0.606 to 0.602).  Among the methods of Table~\ref{tab:results}, \system{} moves from fifth to third on F1, behind Uncertainty (0.726) and PF-MA (0.715), and stays first on AULC-PosR.  Its F1 gap to Uncertainty halves, from 0.144 to 0.073.  At 10\% noise F1 rises from 0.197 to 0.276, and \system{} stays ahead of every baseline, while AULC-PosR falls from 0.464 to 0.447.  In these runs, which ran many tasks at once, selection time rises from 4.05 to 4.29\,s per round.  The main tables keep $K{=}512$, so every dataset uses the same settings.

\section{The Evaluation Classifier on UC~Merced}\label{app:evalclf}

UC~Merced's working set holds 1,680 images, fewer than the 2,000 presumed negatives the evaluation classifier asks for, so it takes every unlabeled item as a negative, including the positives a method has not found.  We re-evaluated the ten methods on UC~Merced with the labeled items alone.  The selection runs are the same.  Only the evaluation classifier differs, so the discovery columns do not change.  Table~\ref{tab:evalclf} gives the values at the budget.  Methods that find nearly every positive keep their F1 within 0.015.  The methods that leave positives unfound gain 0.19 to 0.24 (SEALS, 0.708 to 0.950), and Random gains 0.58.  Retrieval-only, which finds 84\% of the positives, loses F1 (0.871 to 0.821).  With these F1 values on UC~Merced, \system{} keeps the best mean rank over the five datasets (2.50).  We did not re-evaluate the other datasets.  Their working sets are larger, but a larger pool alone does not make this contamination negligible.  The presumed negatives are drawn uniformly from the unlabeled pool, so the expected share of positives among them equals the share of unfound positives in that pool.  We did not measure the resulting effect on F1 on these datasets.

\begin{table}[!ht]
\centering
\caption{UC~Merced, DINOv2, clean labels: values at the budget with the paper's evaluation classifier (labeled items and presumed negatives) and with the labeled items only.  Rerun of the evaluation on a second machine, so the paper's column can differ from Table~\ref{tab:results} in the third decimal.  AULC-PosR and Pos.~Rec.\ depend only on the selection, so they are the same under both classifiers and are listed once.  $^{\dagger}$~marks ablations of \system{}.}
\label{tab:evalclf}
\footnotesize
\setlength{\tabcolsep}{4pt}
\begin{tabular}{l cc cc c cc c}
\toprule
 & \multicolumn{2}{c}{F1} & \multicolumn{2}{c}{Recall} & AULC-PosR & \multicolumn{2}{c}{AULC-F1} & Pos.~Rec. \\
\cmidrule(lr){2-3}\cmidrule(lr){4-5}\cmidrule(lr){6-6}\cmidrule(lr){7-8}\cmidrule(lr){9-9}
Strategy & paper & labeled only & paper & labeled only & both & paper & labeled only & both \\
\midrule
\system{} (ours) & 0.952 & 0.951 & 0.964 & 0.945 & 0.805 & 0.841 & 0.855 & 0.993 \\
SEALS (most-likely-pos.) & 0.948 & 0.934 & 0.958 & 0.947 & 0.824 & 0.853 & 0.802 & 0.990 \\
PF-MA & 0.950 & 0.951 & 0.961 & 0.948 & 0.721 & 0.775 & 0.881 & 0.992 \\
SEALS & 0.708 & 0.950 & 0.614 & 0.946 & 0.307 & 0.494 & 0.880 & 0.514 \\
AnchorAL & 0.724 & 0.944 & 0.637 & 0.947 & 0.348 & 0.526 & 0.815 & 0.546 \\
Uncertainty & 0.765 & 0.953 & 0.702 & 0.945 & 0.409 & 0.594 & 0.907 & 0.628 \\
Random & 0.301 & 0.883 & 0.190 & 0.876 & 0.086 & 0.167 & 0.793 & 0.152 \\
Funnel-RankNet$^{\dagger}$ & 0.952 & 0.950 & 0.965 & 0.945 & 0.805 & 0.842 & 0.854 & 0.995 \\
Funnel-QBC$^{\dagger}$ & 0.733 & 0.954 & 0.639 & 0.949 & 0.432 & 0.632 & 0.884 & 0.550 \\
Retrieval-only$^{\dagger}$ & 0.871 & 0.821 & 0.826 & 0.940 & 0.700 & 0.774 & 0.627 & 0.841 \\
\bottomrule
\end{tabular}
\end{table}

\section{The Trigger's Window}\label{app:trigger-window}

As run, the trigger's window holds the batches of the exploitation phase, including the batch selected before RankNet can train when the warm-up batch is all positive (Section~\ref{sec:method:adaptive}).  That batch has a precision near zero, so it can bring the switch forward.  We ran a variant whose window skips it, counts the warm-up batch instead, and is read only once RankNet can train, on the three benchmarks with clean labels and at 10\% noise.  Paired per task, every column moved by at most 0.006 on every dataset and condition.  The share of tasks entering the mixed phase changed (UC~Merced, clean: 92\% to 63\%), and the switch came about one round earlier where the warm-up batch held a negative.  \system{}'s positions among the external methods did not change, except on CUB-200 with clean labels, where it moved from second to third in F1 and Recall, both near-ties (F1 0.878 to 0.877, Recall 0.889 to 0.886).  Every other result in the paper uses the original rule of Section~\ref{sec:method:adaptive}, not this variant.

\section{Retrieving from the Unlabeled Pool Only}\label{app:unlabeled-pool}

In our retrieval stage the index holds the whole working set, and the labeled images a query finds are dropped afterwards, so a query contributes fewer than $K$ candidates.  We reran \system{}, Funnel-RankNet, Funnel-QBC and Retrieval-only with this retrieval stage on the three benchmarks (DINOv2, clean labels), each query now taking its $K$ nearest \emph{unlabeled} neighbors.  The candidate set grows by 6 to 10\% on average, and by 11 to 25\% for Retrieval-only.  For \system{} every column changed by at most 0.003.  The largest change for any method is Retrieval-only on UC~Merced (F1 $+0.011$, Pos.~Rec.\ $+0.016$).  The methods that rank the candidates with RankNet or the committee lose 0.001 to 0.002 of AULC-PosR on CUB-200.  No position in F1 or AULC-PosR changes outside ties at the third decimal.

\section{Baseline Variants: GAL and PF-MA's Probability Map}\label{app:gal}

\emph{GAL.}  We ran GAL~\citep{bar2024gal} on the three benchmarks, DINOv2, clean and 10\% noise, at batch 20 like every other method, and on UC~Merced also at batch 3.  Its impact scoring runs on up to 200 candidates, the unlabeled items with the highest decision values of a linear SVM refitted on the labeled set each round, following the paper's $K{=}200$ setting.  GAL's selection times include this step.  The paper's main experiments use batches of 1 and 3, and it prefers small batches, because a larger batch is more exposed to wrong pseudo-labels, so batch 20 is the largest departure from its setting.  GAL refits a linear SVM twice for every candidate at every batch slot, which costs 34\,s per round on FGVC-Aircraft (Table~\ref{tab:cost}) and rules out the large corpora.  Values at the budget, GAL (batch 20) against \system{}, F1 / AULC-PosR / Pos.~Rec.: UC~Merced 0.907 / 0.543 / 0.853 against 0.951 / 0.805 / 0.993, CUB-200 0.873 / 0.519 / 0.829 against 0.878 / 0.746 / 0.939, FGVC-Aircraft 0.720 / 0.459 / 0.797 against 0.727 / 0.564 / 0.838.  At 10\% noise: 0.658 / 0.435 / 0.656, 0.673 / 0.383 / 0.635 and 0.512 / 0.336 / 0.589, against 0.839 / 0.696 / 0.878, 0.721 / 0.621 / 0.790 and 0.533 / 0.426 / 0.638.  \system{} is ahead of GAL in every cell, significantly per class except F1 on FGVC-Aircraft, clean ($p{=}0.10$) and at 10\% noise ($p{=}0.004$).  On FGVC-Aircraft GAL has the highest F1 of the external methods in both conditions.  At batch 3 on UC~Merced, with clean labels, GAL gains 0.01 to 0.02 in F1, Recall and Pos.~Rec.\ (F1 0.919, Pos.~Rec.\ 0.871).  Its AULC values depend on the reading, because the batch-3 curve starts at 5 labels and the batch-20 curve at 22.  From each curve's first point, batch 3 has the lower AULC-PosR (0.526 against 0.543).  Over the common interval of 22 to 242 labels it has the higher (0.561).  At 10\% noise batch 3 gains F1 and Recall and loses Pos.~Rec.\ (0.650 against 0.656).  Under either reading GAL stays behind \system{} on AULC-PosR in all 21 classes.  Its selection time per task is the same at both batch sizes, since its cost is per label.

\emph{PF-MA's probability map.}  The PF-MA paper does not state how SVM decision values become probabilities.  Besides the sigmoid reading used in the tables, we ran a Platt-calibrated reading fitted on the labeled set.  With clean labels and DINOv2 the two are close (AULC-PosR within 0.011, F1 within 0.008 on every dataset).  At 10\% noise the Platt reading has the lower AULC-PosR on every dataset with both encoders (by 0.024 to 0.139), so the tables report the sigmoid reading.

\section{The Ablations at 10\% Noise}\label{app:noise-ablations}

Table~\ref{tab:noise-ablations} gives \system{} and its three ablations at 10\% label noise with DINOv2, the rows that Table~\ref{tab:meanrank} and Figure~\ref{fig:noise} use next to Table~\ref{tab:noise}.  Funnel-RankNet differs significantly from \system{} only in AULC-F1 on iNaturalist, where \system{} is higher.  On iNaturalist Funnel-QBC has the highest F1 (0.224 against 0.197, not significant, $p{=}0.22$), with a much lower AULC-PosR (0.211 against 0.464).

\begin{table}[!ht]
\centering
\caption{\system{} and its ablations at 10\% label noise, DINOv2, read as Table~\ref{tab:results}.  Pos.~Rec.\ counts true positives labeled positive.  $^{\dagger}$~marks ablations of \system{}.}
\label{tab:noise-ablations}
\footnotesize
\begin{tabular}{l cccccc}
\toprule
Strategy & F1 & Recall & AULC-PosR & AULC-F1 & Pos.\ Rec. & Labels \\
\midrule
\addlinespace
\multicolumn{7}{l}{\textit{UC Merced (21 classes)}} \\
\system{} (ours) & 0.839 & 0.883 & 0.696 & 0.752 & 0.878 & 202 \\
Funnel-RankNet$^{\dagger}$ & \textbf{0.846} & \textbf{0.888} & \textbf{0.702} & \textbf{0.754} & \textbf{0.883} & 199 \\
Funnel-QBC$^{\dagger}$ & 0.622 & 0.631 & 0.390 & 0.550 & 0.534 & 242 \\
Retrieval-only$^{\dagger}$ & 0.668 & 0.700 & 0.597 & 0.659 & 0.694 & 227 \\
\addlinespace
\multicolumn{7}{l}{\textit{CUB-200-2011 (200 classes)}} \\
\system{} (ours) & \textbf{0.721} & 0.798 & 0.621 & \textbf{0.710} & 0.790 & 138 \\
Funnel-RankNet$^{\dagger}$ & 0.714 & \textbf{0.800} & 0.621 & 0.708 & \textbf{0.791} & 138 \\
Funnel-QBC$^{\dagger}$ & 0.610 & 0.731 & 0.384 & 0.606 & 0.488 & 143 \\
Retrieval-only$^{\dagger}$ & 0.665 & 0.792 & \textbf{0.651} & 0.703 & 0.761 & 136 \\
\addlinespace
\multicolumn{7}{l}{\textit{FGVC-Aircraft (90 classes)}} \\
\system{} (ours) & \textbf{0.533} & \textbf{0.601} & 0.426 & \textbf{0.447} & \textbf{0.638} & 264 \\
Funnel-RankNet$^{\dagger}$ & 0.524 & 0.599 & \textbf{0.427} & 0.444 & 0.636 & 263 \\
Funnel-QBC$^{\dagger}$ & 0.453 & 0.529 & 0.240 & 0.377 & 0.393 & 269 \\
Retrieval-only$^{\dagger}$ & 0.318 & 0.362 & 0.250 & 0.306 & 0.336 & 269 \\
\addlinespace
\multicolumn{7}{l}{\textit{iNaturalist 2021 (20 species)}} \\
\system{} (ours) & 0.197 & 0.604 & 0.464 & \textbf{0.311} & 0.653 & 602 \\
Funnel-RankNet$^{\dagger}$ & 0.145 & \textbf{0.620} & \textbf{0.480} & 0.268 & \textbf{0.670} & 602 \\
Funnel-QBC$^{\dagger}$ & \textbf{0.224} & 0.562 & 0.211 & 0.261 & 0.350 & 602 \\
Retrieval-only$^{\dagger}$ & 0.040 & 0.550 & 0.331 & 0.093 & 0.456 & 602 \\
\addlinespace
\multicolumn{7}{l}{\textit{fMoW (10 categories)}} \\
\system{} (ours) & \textbf{0.343} & 0.322 & 0.174 & 0.263 & 0.309 & 1022 \\
Funnel-RankNet$^{\dagger}$ & \textbf{0.343} & \textbf{0.337} & \textbf{0.189} & 0.263 & \textbf{0.330} & 1022 \\
Funnel-QBC$^{\dagger}$ & 0.317 & 0.332 & 0.101 & \textbf{0.271} & 0.179 & 1022 \\
Retrieval-only$^{\dagger}$ & 0.203 & 0.227 & 0.084 & 0.172 & 0.142 & 1022 \\
\bottomrule
\end{tabular}
\end{table}

\section{Sensitivity of \system{}'s Settings}\label{app:sensitivity}

We changed one setting at a time around the defaults on CUB-200, FGVC-Aircraft and iNaturalist, with DINOv2, clean labels and 10\% noise, and 5 repeats.  CUB-200 and FGVC-Aircraft use all their classes (200 and 90), and iNaturalist its 20 target species.  The settings are the trigger threshold $\tau$ (0.5, 0.6, 0.8, 0.9), the window $w$ (1, 5), the RankNet share of a mixed batch (0.4, 0.8) and the growth factor $\gamma$ (1.0, 1.5).  Tables~\ref{tab:sens-clean} and~\ref{tab:sens-noise} give F1 and AULC-PosR at the budget.  With clean labels every change moves F1 and AULC-PosR by at most 0.012.  The largest are $\gamma{=}1.5$ on CUB-200 (AULC-PosR $-0.012$), $w{=}1$ on CUB-200 (AULC-PosR $-0.009$) and share 0.8 on iNaturalist (F1 $-0.009$).  At 10\% noise the RankNet share matters on iNaturalist: a larger share raises AULC-PosR and lowers F1 (share 0.8: AULC-PosR $+0.011$, F1 $-0.037$), and a smaller one does the reverse (share 0.4: F1 $+0.023$, AULC-PosR $-0.012$).  The share also moves FGVC-Aircraft by up to 0.010.  Every other change stays within 0.008.

\begin{table}[!ht]
\centering
\caption{One-at-a-time sensitivity, DINOv2, clean labels, values at the budget.}
\label{tab:sens-clean}
\footnotesize
\begin{tabular}{l cc cc cc}
\toprule
 & \multicolumn{2}{c}{CUB-200} & \multicolumn{2}{c}{FGVC-Aircraft} & \multicolumn{2}{c}{iNaturalist} \\
\cmidrule(lr){2-3}\cmidrule(lr){4-5}\cmidrule(lr){6-7}
Setting & F1 & AULC-PosR & F1 & AULC-PosR & F1 & AULC-PosR \\
\midrule
default ($\tau{=}0.7$, $w{=}3$, share 0.6, $\gamma{=}1.2$) & 0.878 & 0.746 & 0.727 & 0.564 & 0.581 & 0.606 \\
$\tau{=}0.5$ & 0.878 & 0.745 & 0.727 & 0.564 & 0.579 & 0.610 \\
$\tau{=}0.6$ & 0.878 & 0.746 & 0.728 & 0.564 & 0.576 & 0.608 \\
$\tau{=}0.8$ & 0.878 & 0.746 & 0.727 & 0.563 & 0.581 & 0.604 \\
$\tau{=}0.9$ & 0.878 & 0.746 & 0.727 & 0.563 & 0.583 & 0.603 \\
$w{=}1$ & 0.877 & 0.737 & 0.723 & 0.559 & 0.580 & 0.604 \\
$w{=}5$ & 0.878 & 0.745 & 0.727 & 0.563 & 0.584 & 0.608 \\
RankNet share 0.4 & 0.877 & 0.745 & 0.725 & 0.559 & 0.581 & 0.601 \\
RankNet share 0.8 & 0.878 & 0.746 & 0.724 & 0.565 & 0.572 & 0.611 \\
$\gamma{=}1.0$ (no growth) & 0.879 & 0.752 & 0.727 & 0.565 & 0.579 & 0.608 \\
$\gamma{=}1.5$ & 0.878 & 0.734 & 0.727 & 0.561 & 0.577 & 0.604 \\
\bottomrule
\end{tabular}
\end{table}

\begin{table}[!ht]
\centering
\caption{One-at-a-time sensitivity, DINOv2, 10\% label noise, values at the budget.}
\label{tab:sens-noise}
\footnotesize
\begin{tabular}{l cc cc cc}
\toprule
 & \multicolumn{2}{c}{CUB-200} & \multicolumn{2}{c}{FGVC-Aircraft} & \multicolumn{2}{c}{iNaturalist} \\
\cmidrule(lr){2-3}\cmidrule(lr){4-5}\cmidrule(lr){6-7}
Setting & F1 & AULC-PosR & F1 & AULC-PosR & F1 & AULC-PosR \\
\midrule
default ($\tau{=}0.7$, $w{=}3$, share 0.6, $\gamma{=}1.2$) & 0.721 & 0.621 & 0.533 & 0.426 & 0.197 & 0.464 \\
$\tau{=}0.5$ & 0.721 & 0.621 & 0.536 & 0.426 & 0.198 & 0.468 \\
$\tau{=}0.6$ & 0.721 & 0.621 & 0.531 & 0.424 & 0.199 & 0.465 \\
$\tau{=}0.8$ & 0.721 & 0.621 & 0.532 & 0.425 & 0.197 & 0.464 \\
$\tau{=}0.9$ & 0.721 & 0.621 & 0.532 & 0.425 & 0.197 & 0.464 \\
$w{=}1$ & 0.724 & 0.616 & 0.536 & 0.422 & 0.190 & 0.461 \\
$w{=}5$ & 0.717 & 0.621 & 0.532 & 0.426 & 0.203 & 0.464 \\
RankNet share 0.4 & 0.723 & 0.619 & 0.529 & 0.416 & 0.220 & 0.452 \\
RankNet share 0.8 & 0.715 & 0.621 & 0.530 & 0.426 & 0.160 & 0.475 \\
$\gamma{=}1.0$ (no growth) & 0.722 & 0.621 & 0.532 & 0.426 & 0.197 & 0.464 \\
$\gamma{=}1.5$ & 0.722 & 0.620 & 0.532 & 0.426 & 0.196 & 0.464 \\
\bottomrule
\end{tabular}
\end{table}

\section{The Second Encoder in Full}\label{app:clip}

Tables~\ref{tab:clip-clean} to~\ref{tab:clip-rank} repeat Tables~\ref{tab:results}, \ref{tab:ablations}, \ref{tab:noise}, \ref{tab:noise-ablations} and~\ref{tab:meanrank} with CLIP ViT-B/16~\citep{radford2021clip}, under the identical protocol.

\begin{table}[!ht]
\centering
\caption{External methods, CLIP ViT-B/16, clean labels, read as Table~\ref{tab:results}.}
\label{tab:clip-clean}
\footnotesize
\begin{tabular}{l cccccc}
\toprule
Strategy & F1 & Recall & AULC-PosR & AULC-F1 & Pos.\ Rec. & Labels \\
\midrule
\addlinespace
\multicolumn{7}{l}{\textit{UC Merced (21 classes)}} \\
\system{} (ours) & \textbf{0.957} & \textbf{0.971} & \textbf{0.810} & \textbf{0.843} & \textbf{0.997} & 171 \\
SEALS (most-likely-pos.) & 0.953 & 0.968 & 0.802 & 0.833 & 0.994 & 187 \\
PF-MA & 0.956 & \textbf{0.971} & 0.726 & 0.777 & 0.996 & 175 \\
SEALS & 0.626 & 0.520 & 0.275 & 0.434 & 0.460 & 242 \\
AnchorAL & 0.650 & 0.550 & 0.311 & 0.471 & 0.496 & 242 \\
Uncertainty & 0.746 & 0.677 & 0.408 & 0.569 & 0.629 & 242 \\
Random & 0.305 & 0.193 & 0.086 & 0.166 & 0.152 & 242 \\
\addlinespace
\multicolumn{7}{l}{\textit{CUB-200-2011 (200 classes)}} \\
\system{} (ours) & \textbf{0.754} & \textbf{0.756} & \textbf{0.633} & \textbf{0.669} & \textbf{0.872} & 139 \\
SEALS (most-likely-pos.) & 0.746 & 0.755 & 0.571 & 0.630 & 0.844 & 141 \\
PF-MA & 0.744 & 0.747 & 0.496 & 0.567 & 0.844 & 141 \\
SEALS & 0.639 & 0.596 & 0.196 & 0.422 & 0.365 & 143 \\
AnchorAL & 0.648 & 0.617 & 0.244 & 0.497 & 0.412 & 143 \\
Uncertainty & 0.717 & 0.711 & 0.333 & 0.523 & 0.628 & 143 \\
Random & 0.067 & 0.041 & 0.030 & 0.047 & 0.036 & 143 \\
\addlinespace
\multicolumn{7}{l}{\textit{FGVC-Aircraft (90 classes)}} \\
\system{} (ours) & 0.465 & 0.446 & \textbf{0.403} & \textbf{0.371} & \textbf{0.616} & 262 \\
SEALS (most-likely-pos.) & 0.463 & 0.442 & 0.371 & 0.356 & 0.580 & 264 \\
PF-MA & \textbf{0.467} & \textbf{0.447} & 0.348 & 0.341 & 0.591 & 264 \\
SEALS & 0.403 & 0.364 & 0.171 & 0.260 & 0.343 & 269 \\
AnchorAL & 0.408 & 0.368 & 0.198 & 0.282 & 0.363 & 269 \\
Uncertainty & 0.450 & 0.423 & 0.281 & 0.324 & 0.520 & 269 \\
Random & 0.087 & 0.053 & 0.029 & 0.051 & 0.044 & 269 \\
\addlinespace
\multicolumn{7}{l}{\textit{iNaturalist 2021 (20 species)}} \\
\system{} (ours) & 0.361 & 0.476 & \textbf{0.424} & 0.350 & 0.629 & 602 \\
SEALS (most-likely-pos.) & 0.364 & \textbf{0.490} & 0.382 & 0.301 & 0.611 & 602 \\
PF-MA & 0.457 & 0.474 & 0.389 & 0.352 & \textbf{0.648} & 602 \\
SEALS & 0.384 & 0.468 & 0.095 & 0.285 & 0.201 & 602 \\
AnchorAL & 0.301 & 0.470 & 0.131 & 0.248 & 0.249 & 602 \\
Uncertainty & \textbf{0.517} & 0.471 & 0.279 & \textbf{0.401} & 0.518 & 602 \\
Random & 0.071 & 0.058 & 0.005 & 0.070 & 0.005 & 602 \\
\addlinespace
\multicolumn{7}{l}{\textit{fMoW (10 categories)}} \\
\system{} (ours) & 0.428 & 0.373 & \textbf{0.225} & 0.341 & 0.397 & 1022 \\
SEALS (most-likely-pos.) & 0.430 & 0.387 & 0.221 & 0.337 & \textbf{0.405} & 1022 \\
PF-MA & \textbf{0.443} & \textbf{0.405} & 0.187 & \textbf{0.354} & 0.385 & 1022 \\
SEALS & 0.389 & 0.392 & 0.072 & 0.295 & 0.142 & 1022 \\
AnchorAL & 0.348 & 0.342 & 0.078 & 0.270 & 0.145 & 1022 \\
Uncertainty & \textbf{0.443} & 0.385 & 0.147 & \textbf{0.354} & 0.285 & 1022 \\
Random & 0.197 & 0.155 & 0.012 & 0.113 & 0.020 & 1022 \\
\bottomrule
\end{tabular}
\end{table}

\begin{table}[!ht]
\centering
\caption{\system{} and its ablations, CLIP ViT-B/16, clean labels.  $^{\dagger}$~marks ablations of \system{}.}
\label{tab:clip-abl-clean}
\footnotesize
\begin{tabular}{l cccccc}
\toprule
Strategy & F1 & Recall & AULC-PosR & AULC-F1 & Pos.\ Rec. & Labels \\
\midrule
\addlinespace
\multicolumn{7}{l}{\textit{UC Merced (21 classes)}} \\
\system{} (ours) & \textbf{0.957} & 0.971 & \textbf{0.810} & \textbf{0.843} & \textbf{0.997} & 171 \\
Funnel-RankNet$^{\dagger}$ & \textbf{0.957} & \textbf{0.972} & 0.809 & 0.842 & \textbf{0.997} & 175 \\
Funnel-QBC$^{\dagger}$ & 0.695 & 0.598 & 0.417 & 0.602 & 0.527 & 242 \\
Retrieval-only$^{\dagger}$ & 0.894 & 0.858 & 0.669 & 0.778 & 0.854 & 241 \\
\addlinespace
\multicolumn{7}{l}{\textit{CUB-200-2011 (200 classes)}} \\
\system{} (ours) & \textbf{0.754} & 0.756 & 0.633 & \textbf{0.669} & 0.872 & 139 \\
Funnel-RankNet$^{\dagger}$ & 0.752 & \textbf{0.760} & \textbf{0.634} & \textbf{0.669} & \textbf{0.876} & 139 \\
Funnel-QBC$^{\dagger}$ & 0.721 & 0.711 & 0.403 & 0.619 & 0.621 & 143 \\
Retrieval-only$^{\dagger}$ & 0.701 & 0.685 & 0.501 & 0.644 & 0.641 & 143 \\
\addlinespace
\multicolumn{7}{l}{\textit{FGVC-Aircraft (90 classes)}} \\
\system{} (ours) & 0.465 & 0.446 & \textbf{0.403} & \textbf{0.371} & \textbf{0.616} & 262 \\
Funnel-RankNet$^{\dagger}$ & \textbf{0.466} & \textbf{0.449} & \textbf{0.403} & 0.370 & 0.615 & 262 \\
Funnel-QBC$^{\dagger}$ & 0.437 & 0.408 & 0.302 & 0.349 & 0.495 & 269 \\
Retrieval-only$^{\dagger}$ & 0.392 & 0.352 & 0.277 & 0.319 & 0.394 & 269 \\
\addlinespace
\multicolumn{7}{l}{\textit{iNaturalist 2021 (20 species)}} \\
\system{} (ours) & 0.361 & 0.476 & 0.424 & 0.350 & 0.629 & 602 \\
Funnel-RankNet$^{\dagger}$ & 0.335 & \textbf{0.489} & \textbf{0.428} & 0.330 & \textbf{0.632} & 602 \\
Funnel-QBC$^{\dagger}$ & \textbf{0.371} & 0.475 & 0.304 & \textbf{0.357} & 0.501 & 602 \\
Retrieval-only$^{\dagger}$ & 0.205 & 0.438 & 0.219 & 0.188 & 0.338 & 602 \\
\addlinespace
\multicolumn{7}{l}{\textit{fMoW (10 categories)}} \\
\system{} (ours) & 0.428 & 0.373 & 0.225 & \textbf{0.341} & 0.397 & 1022 \\
Funnel-RankNet$^{\dagger}$ & \textbf{0.429} & \textbf{0.388} & \textbf{0.228} & 0.338 & \textbf{0.408} & 1022 \\
Funnel-QBC$^{\dagger}$ & 0.418 & 0.380 & 0.140 & 0.339 & 0.254 & 1022 \\
Retrieval-only$^{\dagger}$ & 0.330 & 0.278 & 0.105 & 0.258 & 0.191 & 1022 \\
\bottomrule
\end{tabular}
\end{table}

\begin{table}[!ht]
\centering
\caption{External methods, CLIP ViT-B/16, 10\% label noise.}
\label{tab:clip-noise}
\footnotesize
\begin{tabular}{l cccccc}
\toprule
Strategy & F1 & Recall & AULC-PosR & AULC-F1 & Pos.\ Rec. & Labels \\
\midrule
\addlinespace
\multicolumn{7}{l}{\textit{UC Merced (21 classes)}} \\
\system{} (ours) & \textbf{0.850} & \textbf{0.899} & \textbf{0.710} & \textbf{0.764} & \textbf{0.892} & 196 \\
SEALS (most-likely-pos.) & 0.844 & 0.881 & 0.708 & 0.757 & 0.891 & 192 \\
PF-MA & 0.837 & 0.887 & 0.618 & 0.666 & 0.887 & 194 \\
SEALS & 0.235 & 0.200 & 0.105 & 0.176 & 0.158 & 242 \\
AnchorAL & 0.355 & 0.329 & 0.182 & 0.298 & 0.253 & 242 \\
Uncertainty & 0.512 & 0.494 & 0.259 & 0.369 & 0.435 & 242 \\
Random & 0.217 & 0.190 & 0.081 & 0.145 & 0.141 & 242 \\
\addlinespace
\multicolumn{7}{l}{\textit{CUB-200-2011 (200 classes)}} \\
\system{} (ours) & \textbf{0.581} & \textbf{0.640} & \textbf{0.498} & \textbf{0.536} & \textbf{0.698} & 142 \\
SEALS (most-likely-pos.) & 0.527 & 0.622 & 0.446 & 0.466 & 0.666 & 141 \\
PF-MA & 0.422 & 0.496 & 0.272 & 0.293 & 0.529 & 143 \\
SEALS & 0.155 & 0.205 & 0.057 & 0.115 & 0.089 & 143 \\
AnchorAL & 0.292 & 0.364 & 0.137 & 0.262 & 0.198 & 143 \\
Uncertainty & 0.278 & 0.318 & 0.099 & 0.179 & 0.195 & 143 \\
Random & 0.045 & 0.061 & 0.029 & 0.043 & 0.035 & 143 \\
\addlinespace
\multicolumn{7}{l}{\textit{FGVC-Aircraft (90 classes)}} \\
\system{} (ours) & \textbf{0.339} & 0.356 & \textbf{0.304} & \textbf{0.289} & \textbf{0.451} & 265 \\
SEALS (most-likely-pos.) & 0.332 & \textbf{0.372} & 0.290 & 0.274 & 0.444 & 265 \\
PF-MA & 0.297 & 0.327 & 0.229 & 0.221 & 0.398 & 265 \\
SEALS & 0.127 & 0.160 & 0.055 & 0.089 & 0.096 & 269 \\
AnchorAL & 0.181 & 0.208 & 0.099 & 0.146 & 0.162 & 269 \\
Uncertainty & 0.233 & 0.244 & 0.111 & 0.156 & 0.222 & 269 \\
Random & 0.049 & 0.075 & 0.028 & 0.038 & 0.041 & 269 \\
\addlinespace
\multicolumn{7}{l}{\textit{iNaturalist 2021 (20 species)}} \\
\system{} (ours) & \textbf{0.132} & \textbf{0.368} & \textbf{0.278} & \textbf{0.154} & \textbf{0.437} & 602 \\
SEALS (most-likely-pos.) & 0.049 & 0.283 & 0.197 & 0.066 & 0.317 & 602 \\
PF-MA & 0.052 & 0.205 & 0.124 & 0.043 & 0.234 & 602 \\
SEALS & 0.006 & 0.148 & 0.009 & 0.010 & 0.012 & 602 \\
AnchorAL & 0.023 & 0.278 & 0.043 & 0.035 & 0.068 & 602 \\
Uncertainty & 0.039 & 0.059 & 0.016 & 0.023 & 0.034 & 602 \\
Random & 0.000 & 0.127 & 0.005 & 0.003 & 0.005 & 602 \\
\addlinespace
\multicolumn{7}{l}{\textit{fMoW (10 categories)}} \\
\system{} (ours) & \textbf{0.345} & 0.327 & \textbf{0.168} & \textbf{0.263} & \textbf{0.306} & 1022 \\
SEALS (most-likely-pos.) & 0.323 & 0.335 & 0.164 & 0.252 & \textbf{0.306} & 1022 \\
PF-MA & 0.329 & \textbf{0.355} & 0.133 & 0.261 & 0.279 & 1022 \\
SEALS & 0.162 & 0.278 & 0.025 & 0.132 & 0.045 & 1022 \\
AnchorAL & 0.192 & 0.253 & 0.045 & 0.161 & 0.078 & 1022 \\
Uncertainty & 0.293 & 0.272 & 0.067 & 0.216 & 0.144 & 1022 \\
Random & 0.065 & 0.240 & 0.011 & 0.058 & 0.019 & 1022 \\
\bottomrule
\end{tabular}
\end{table}

\begin{table}[!ht]
\centering
\caption{\system{} and its ablations, CLIP ViT-B/16, 10\% label noise.  $^{\dagger}$~marks ablations of \system{}.}
\label{tab:clip-abl-noise}
\footnotesize
\begin{tabular}{l cccccc}
\toprule
Strategy & F1 & Recall & AULC-PosR & AULC-F1 & Pos.\ Rec. & Labels \\
\midrule
\addlinespace
\multicolumn{7}{l}{\textit{UC Merced (21 classes)}} \\
\system{} (ours) & \textbf{0.850} & \textbf{0.899} & 0.710 & 0.764 & 0.892 & 196 \\
Funnel-RankNet$^{\dagger}$ & 0.849 & 0.896 & \textbf{0.715} & \textbf{0.765} & \textbf{0.894} & 192 \\
Funnel-QBC$^{\dagger}$ & 0.592 & 0.587 & 0.365 & 0.518 & 0.507 & 242 \\
Retrieval-only$^{\dagger}$ & 0.669 & 0.714 & 0.552 & 0.641 & 0.684 & 242 \\
\addlinespace
\multicolumn{7}{l}{\textit{CUB-200-2011 (200 classes)}} \\
\system{} (ours) & 0.581 & 0.640 & 0.498 & 0.536 & 0.698 & 142 \\
Funnel-RankNet$^{\dagger}$ & \textbf{0.583} & \textbf{0.656} & \textbf{0.503} & \textbf{0.537} & \textbf{0.713} & 141 \\
Funnel-QBC$^{\dagger}$ & 0.463 & 0.541 & 0.285 & 0.431 & 0.407 & 143 \\
Retrieval-only$^{\dagger}$ & 0.449 & 0.552 & 0.400 & 0.482 & 0.493 & 143 \\
\addlinespace
\multicolumn{7}{l}{\textit{FGVC-Aircraft (90 classes)}} \\
\system{} (ours) & \textbf{0.339} & 0.356 & 0.304 & 0.289 & 0.451 & 265 \\
Funnel-RankNet$^{\dagger}$ & 0.338 & \textbf{0.365} & \textbf{0.309} & \textbf{0.290} & \textbf{0.458} & 265 \\
Funnel-QBC$^{\dagger}$ & 0.299 & 0.315 & 0.204 & 0.253 & 0.327 & 269 \\
Retrieval-only$^{\dagger}$ & 0.245 & 0.282 & 0.221 & 0.241 & 0.298 & 269 \\
\addlinespace
\multicolumn{7}{l}{\textit{iNaturalist 2021 (20 species)}} \\
\system{} (ours) & 0.132 & 0.368 & 0.278 & \textbf{0.154} & 0.437 & 602 \\
Funnel-RankNet$^{\dagger}$ & 0.096 & \textbf{0.387} & \textbf{0.296} & 0.136 & \textbf{0.459} & 602 \\
Funnel-QBC$^{\dagger}$ & \textbf{0.135} & 0.286 & 0.144 & 0.124 & 0.272 & 602 \\
Retrieval-only$^{\dagger}$ & 0.028 & 0.274 & 0.131 & 0.049 & 0.190 & 602 \\
\addlinespace
\multicolumn{7}{l}{\textit{fMoW (10 categories)}} \\
\system{} (ours) & \textbf{0.345} & 0.327 & 0.168 & \textbf{0.263} & 0.306 & 1022 \\
Funnel-RankNet$^{\dagger}$ & 0.335 & \textbf{0.330} & \textbf{0.175} & 0.259 & \textbf{0.312} & 1022 \\
Funnel-QBC$^{\dagger}$ & 0.313 & 0.299 & 0.096 & 0.249 & 0.182 & 1022 \\
Retrieval-only$^{\dagger}$ & 0.185 & 0.195 & 0.072 & 0.161 & 0.122 & 1022 \\
\bottomrule
\end{tabular}
\end{table}

\begin{table}[!ht]
\centering
\caption{Mean rank of the ten methods, CLIP ViT-B/16, read as Table~\ref{tab:meanrank} (ranks use unrounded values).  $^{\dagger}$~marks ablations of \system{}.}
\label{tab:clip-rank}
\footnotesize
\begin{tabular}{l cccc cccc}
\toprule
 & \multicolumn{4}{c}{Clean labels} & \multicolumn{4}{c}{10\% label noise} \\
\cmidrule(lr){2-5}\cmidrule(lr){6-9}
Strategy & Pos. & Discovery & F1 & Overall & Pos. & Discovery & F1 & Overall \\
\midrule
\system{} (ours) & 1 & 1.60 & 3.20 & 2.40 & 2 & 2.00 & 1.40 & 1.70 \\
Funnel-RankNet$^{\dagger}$ & 1 & 1.40 & 3.40 & 2.40 & 1 & 1.00 & 2.00 & 1.50 \\
PF-MA & 3 & 4.00 & 2.40 & 3.20 & 4 & 4.80 & 4.40 & 4.60 \\
SEALS (most-likely-pos.) & 4 & 3.20 & 3.80 & 3.50 & 3 & 3.00 & 3.60 & 3.30 \\
Uncertainty & 5 & 6.20 & 3.80 & 5.00 & 7 & 7.40 & 6.80 & 7.10 \\
Funnel-QBC$^{\dagger}$ & 6 & 5.60 & 5.60 & 5.60 & 4 & 5.20 & 4.00 & 4.60 \\
Retrieval-only$^{\dagger}$ & 7 & 6.00 & 7.80 & 6.90 & 6 & 5.00 & 6.20 & 5.60 \\
AnchorAL & 8 & 8.00 & 7.80 & 7.90 & 8 & 7.60 & 7.60 & 7.60 \\
SEALS & 9 & 9.00 & 7.20 & 8.10 & 9 & 9.00 & 9.00 & 9.00 \\
Random & 10 & 10.00 & 10.00 & 10.00 & 10 & 10.00 & 10.00 & 10.00 \\
\bottomrule
\end{tabular}
\end{table}

\section{Target Classes of the Large Corpora}\label{app:classes}

\begin{table}[!ht]
\centering
\caption{Pre-registered target classes for the large-corpus study.  Counts are on each full corpus (iNaturalist 2{,}686{,}843 images, fMoW 83{,}412), prevalence is $n_{\mathrm{pos}}$ divided by the corpus size, and the 80/20 split leaves 80\% of each class in the work pool.  The twelve iNaturalist species with 300 instances sit at the dataset's per-class cap and come from eleven taxonomic groups (two are plants).  \texttt{space\_facility}'s ${\sim}9$ test positives quantize its test recall in steps of ${\sim}0.11$, which affects only the secondary F1 metric.}
\label{tab:scale-classes}
\footnotesize
\begin{tabular}{r l l}
\toprule
$n_{\mathrm{pos}}$ & Prevalence & Class \\
\midrule
\multicolumn{3}{l}{\textit{iNaturalist 2021 (20 species)}} \\
\addlinespace
157 & 0.0058\% & \textit{Erica corifolia} (Plants) \\
178 & 0.0066\% & \textit{Rosa spinosissima} (Plants) \\
191 & 0.0071\% & \textit{Packera tampicana} (Plants) \\
200 & 0.0074\% & \textit{Brodiaea terrestris} (Plants) \\
223 & 0.0083\% & \textit{Turdus falcklandii} (Birds) \\
233 & 0.0087\% & \textit{Adelphocoris lineolatus} (Insects) \\
262 & 0.0098\% & \textit{Euphorbia fendleri} (Plants) \\
279 & 0.0104\% & \textit{Diaphania hyalinata} (Insects) \\
300 & 0.0112\% & \textit{Bufo bankorensis} (Amphibians) \\
300 & 0.0112\% & \textit{Velella velella} (Animalia) \\
300 & 0.0112\% & \textit{Salticus scenicus} (Arachnids) \\
300 & 0.0112\% & \textit{Aphelocoma wollweberi} (Birds) \\
300 & 0.0112\% & \textit{Crucibulum laeve} (Fungi) \\
300 & 0.0112\% & \textit{Ischnura elegans} (Insects) \\
300 & 0.0112\% & \textit{Ovis canadensis} (Mammals) \\
300 & 0.0112\% & \textit{Arion subfuscus} (Mollusks) \\
300 & 0.0112\% & \textit{Anticlea elegans} (Plants) \\
300 & 0.0112\% & \textit{Centaurea nigra} (Plants) \\
300 & 0.0112\% & \textit{Ostracion meleagris} (Ray-finned Fishes) \\
300 & 0.0112\% & \textit{Varanus bengalensis} (Reptiles) \\
\midrule
\multicolumn{3}{l}{\textit{fMoW (10 categories)}} \\
\addlinespace
44 & 0.053\% & \texttt{space\_facility} \\
544 & 0.652\% & \texttt{flooded\_road} \\
681 & 0.816\% & \texttt{stadium} \\
810 & 0.971\% & \texttt{factory\_or\_powerplant} \\
1{,}177 & 1.411\% & \texttt{storage\_tank} \\
1{,}195 & 1.433\% & \texttt{burial\_site} \\
1{,}217 & 1.459\% & \texttt{race\_track} \\
1{,}853 & 2.222\% & \texttt{tunnel\_opening} \\
2{,}919 & 3.500\% & \texttt{parking\_lot\_or\_garage} \\
4{,}521 & 5.420\% & \texttt{place\_of\_worship} \\
\bottomrule
\end{tabular}
\end{table}

\end{document}